\documentclass[final, 3p, times]{elsarticle}

\makeatletter
\def\ps@pprintTitle{%
\let\@oddhead\@empty
\let\@evenhead\@empty
\def\@oddfoot{}%
\let\@evenfoot\@oddfoot}
\makeatother

\usepackage{hyperref,amsmath,amsthm, cleveref, scalerel, stackengine, pgfplots, dsfont, booktabs, multirow, multicol}
\usepackage[ruled,vlined]{algorithm2e}
\usepackage{xcolor}
\usepackage{caption, subcaption}
\usepackage{bm}
\usepackage[bottom]{footmisc}
\journal{journal name}

\stackMath
\newcommand\reallywidehat[1]{%
\savestack{\tmpbox}{\stretchto{%
  \scaleto{%
    \scalerel*[\widthof{\ensuremath{#1}}]{\kern-.6pt\bigwedge\kern-.6pt}%
    {\rule[-\textheight/2]{1ex}{\textheight}}%WIDTH-LIMITED BIG WEDGE
  }{\textheight}%
}{0.8ex}}%
\stackon[1pt]{#1}{\tmpbox}%
}
\numberwithin{equation}{section}

\begin{document}

\begin{frontmatter}

\title{Forecasting VIX using interpretable Kolmogorov-Arnold networks}

\author[1]{So-Yoon Cho}
\ead{soyooooon@yonsei.ac.kr}

\author[1]{Sungchul Lee}
\address[1]{Department of Mathematics, Yonsei University, Seoul 03722, Republic of Korea}
\ead{sungchul@yonsei.ac.kr}

\author[3]{Hyun-Gyoon Kim\corref{hgkim}}
\address[3]{Department of Financial Engineering, Ajou University, Gyeonggi-do 16499, Republic of Korea}
\ead{hyungyoonkim@ajou.ac.kr}

\cortext[hgkim]{Corresponding author}

\begin{abstract}

This paper presents the use of Kolmogorov-Arnold Networks (KANs) for forecasting the CBOE Volatility Index (VIX). 
Unlike traditional MLP-based neural networks that are often criticized for their black-box nature, KAN offers an interpretable approach via learnable spline-based activation functions and symbolification. 
Based on a parsimonious architecture with symbolic functions, KAN expresses a forecast of the VIX as a closed-form in terms of explanatory variables, and provide interpretable insights into key characteristics of the VIX, including mean reversion and the leverage effect. 
Through in-depth empirical analysis across multiple datasets and periods, we show that KANs achieve competitive forecasting performance while requiring significantly fewer parameters compared to MLP-based neural network models. 
Our findings demonstrate the capacity and potential of KAN as an interpretable financial time-series forecasting method.

\end{abstract}


\end{frontmatter}

% KAN or KANs - 단복수 통일

\section{Introduction}

Deep learning based on the universal approximation theorem and implemented via multi-layer perceptrons (MLPs) has substantially improved over the past few decades and has been applied across a wide range of fields and industries. This progress can be attributed to its advantages in, for example, feature extraction and representation \cite{lecun2015deep}, function approximation \cite{cybenko1989approximation}, and forecasting performance \cite{zhang1998forecasting}.
However, in finance where \emph{interpretability} is crucial for decision-making, the black-box nature of MLPs often makes decision-makers hesitant to fully adopting deep learning. 
In particular, for financial time-series forecasting, the lack of interpretability raises concerns about their robustness to future market dynamics.

In financial time-series data, volatility is regarded as one of the most important measures along with returns.
Among the various measures of volatility, the Chicago Board Options Exchange (CBOE) Volatility Index (VIX) represents an important indicator of expected market fluctuation.
The VIX, often referred to as the ``fear index'', reflects the market's expectations of volatility for the S\&P 500 index over the next 30 days. It is calculated by combining the weighted prices of S\&P 500 index call and put options across various strike prices, thereby providing a forward-looking indicator of market uncertainty. Unlike historical volatility measures, which rely entirely on past data, options-implied volatility measures such as the VIX incorporate the forward-looking views of option market participants. As a result, the VIX often provides practitioners with a more valuable perspective on market risk than historical backward-looking metrics.

Beyond reflecting the market’s expectation of volatility, the VIX also serves as the underlying asset for various tradable products following the launch of VIX derivatives such as VIX futures and VIX options in the 2000s. Notably, the well-documented negative correlation between volatility and stock market returns, often called the leverage effect, makes these volatility-based instruments appealing hedging tools in portfolio management. Consequently, accurate VIX forecasting is critical not only for understanding market participants' expectations but also for preparing for and effectively managing upcoming risks.

Extensive research has been conducted on modeling and forecasting the VIX, building on empirical stylized facts of volatility such as volatility clustering \cite{cont2001empirical}, long-range dependence \cite{baillie1996long}, and mean reversion \cite{taylor2008modelling}. Commonly used traditional approaches in the financial time-series modeling literature include ARMA-type and GARCH-type models \cite{satchell2011forecasting}.
The ARMA model \cite{box2015time} integrates autoregressive (AR) and moving average (MA) components, leveraging past values and errors to effectively describe time series data. Similarly, the ARIMA model \cite{box2015time} extends ARMA by incorporating differencing to address non-stationarity \cite{french1987expected, fernandes2014modeling}.
Another notable method for modeling volatility is the Heterogeneous Autoregressive (HAR) model \cite{corsi2009simple} which is based on the heterogeneous market hypothesis \cite{muller2008fractals}. 
The HAR framework presumes that future volatility depends on averages of past volatility over multiple time horizons (e.g., daily, weekly, monthly) to capture the mean-reverting behavior and long-range dependence \cite{corsi2009simple}.
Empirical findings suggest that the HAR model often exhibits strong forecasting performances, as demonstrated in \cite{andersen2007roughing, patton2009optimal, izzeldin2019forecasting}, where its additive cascade structure aligns well with real-world market data and improves predictive accuracy.

The generalized autoregressive conditional heteroskedasticity (GARCH) models \cite{bollerslev1986generalized} are also widely employed to capture conditional variance in financial time series.
However, direct application of GARCH for the VIX may not be appropriate for two main reasons.
First, GARCH relies on past stock returns to infer future volatility, whereas the VIX is estimated from option prices and thus captures forward-looking expectations of market participants\footnote{Although both GARCH and ARMA-type models depend on historical data, ARMA-type approaches can directly model the VIX itself by using its past values (i.e., past VIX data) to forecast future VIX. By contrast, GARCH’s dependence on past stock returns does not capture the volatility implied by option data.}.
Second, GARCH models are typically defined under the real (physical) probability measure, whereas the VIX is extracted from option, which are priced under the risk-neutral measure.
In other words, GARCH-based estimates and the VIX are not in the same measure space, and therefore cannot be directly comparable.
% cascade structure:
% Each term in the model corresponds to a specific component of market volatility. The short-term volatility is driven by daily trading or high-frequency traders. The mid-term volatility reflects the behavior of participants who rebalance their positions weekly or trade less frequently than short-term traders. Finally, the long-term volatility is influenced by institutional investors, such as pension funds or insurance companies, who engage in infrequent rebalancing. Additionally, the level of long-term volatility plays a crucial role in determining short-term volatility, while the reverse is not true. This hierarchical cascade structure is a fundamental characteristic of the model.
% 

In recent years, deep learning approaches, particularly those based on MLPs and recurrent neural networks (RNNs), have demonstrated notable success in volatility forecasting \cite{ge2022neural}. As noted by \citet{ge2022neural}, many deep learning methods rely on MLPs \cite{mostafa2015computational, kumar2015estimation, pyo2018exploiting} or RNN-based Long Short-Term Memory (LSTM) networks \cite{petnehazi2019exploring, kim2018forecasting, liu2019novel} to predict historical or realized volatility.
For VIX forecasting, \citet{lahmiri2016variational} proposed the VMD-GRNN model, combining variational mode decomposition (VMD) with a general regression neural network (GRNN). Similarly, \citet{litimi2018chaotic} investigated chaotic map-based models and found that neural networks outperformed other methods in out-of-sample VIX forecasting across multiple markets.
Despite these encouraging results, MLP-based models have often been criticized for their black-box nature.
This presents a serious disadvantage particularly in finance as decision-makers may hesitate to trust models whose inner workings are not clearly understood.
Although a model may perform well on historical data, the lack of interpretability raises concerns about its robustness to future market dynamics.
% Thus, while MLP-based deep learning architectures can achieve promising results, their limited interpretability may hinder broader adoption by finance practitioners.

To address these limitations, we adopt Kolmogorov-Arnold Networks (KANs) \cite{liu2024kan} for VIX forecasting. Based on the Kolmogorov-Arnold representation theorem \cite{kolmogorov1957representation, kolmogorov1961representation}, KANs offer a promising alternative to conventional MLPs by employing learnable spline-based univariate functions along each network edge. This structure provides both interpretability and reliable forecasting accuracy.
Specifically, KANs allow researchers to visualize how individual splines and activation functions evolve during training and thereby uncover intricate data patterns. Additionally, KANs can approximate complex target functions with potentially fewer parameters than typical MLP-based deep learning models, making them a more efficient approach to VIX forecasting.

Despite being introduced only in 2024, KANs have already garnered considerable attention across a range of time-series forecasting tasks \cite{peng2024predictive, vaca2024kolmogorov, alves2024use}.
In finance, KANs have been utilized to predict, for example, implied volatility \cite{xu2024kolmogorov}, cryptocurrency absolute returns \cite{inzirillo2024sigkan}, and trading volume \cite{genet2024tkan, inzirillo2024sigkan, genet2024temporal}, and have demonstrated competitive or even superior performance compared to conventional MLP or LSTM models.
Most existing studies, however, integrate KAN modules into more elaborate architectures (e.g., RNNs, convolutional networks, or gated residual networks) to enhance predictive power. 
Although these designs can improve accuracy, they also compromise KAN’s key strength: interpretability. 
In contrast, our aim is to overcome the black-box limitations of MLPs and produce interpretable financial time-series forecasts by employing a pure KAN structure. 
Rather than blending KANs with complicated modules that dilute its transparency, we utilize KAN’s inherent capacity for visual identification and mathematical characterization of time-series features in the data.
% Moreover, expressing forecasts in closed-form composed of symbolic functions helps mitigate potential instabilities when encountering previously unseen data.

% model electrohydrodynamic pump dynamics \cite{peng2024predictive} 
% forecast satellite traffic \cite{vaca2024kolmogorov}
% perform wind nowcasting \cite{alves2024use}. 

% Temporal KANs \cite{genet2024tkan}
% SigKAN \cite{inzirillo2024sigkan}
% attention-based Temporal KAN \cite{genet2024temporal}) 
% Time-series KAN (TKAN) and Multivariate Time-series KAN (MT-KAN) \citet{xu2024kolmogorov} 

The remainder of this paper is organized as follows. \Cref{sec: background of KANs} reviews the Kolmogorov-Arnold representation theorem and describes the structure and training procedure of KANs. \Cref{sec: KANs for vix forecasting} analyzes empirical results based on the interpretability of KAN and evaluates its forecasting performances through comparisons with traditional time-series models and MLP-based neural networks.
\Cref{sec: conclusion} concludes the paper.

\section{Kolmogorov-Arnold Networks: Representation, Structure, and Training} \label{sec: background of KANs}

\subsection{Kolmogorov-Arnold Representation Theorem}
The Kolmogorov-Arnold representation theorem \cite{kolmogorov1957representation, kolmogorov1961representation} provides a foundational framework for expressing complex multivariate functions through compositions of univariate functions and additive operations. Formally, the theorem states that any continuous multivariable function \( f : [0, 1]^n \to \mathbb{R} \) can be represented as a finite sum of compositions of continuous univariate functions: 
\begin{equation}
f(\mathbf{x}) = f(x_1, x_2, \ldots, x_n) = \sum_{q=1}^{2n+1} \Phi_q\left(\sum_{p=1}^n \phi_{q,p}(x_p)\right),
\end{equation}
where \( \phi_{q, p} : [0, 1] \to \mathbb{R} \) and \( \Phi_q : \mathbb{R} \to \mathbb{R} \) are continuous univariate functions. Note that these functions may exhibit non-smooth or fractal-like behavior.

This theorem is often regarded as highly promising in the field of deep learning, as it reduces the challenge of approximating multivariable functions to the optimization of univariate functions. Nevertheless, its original formulation has certain limitations, notably a shallow two-layer structure and restricted flexibility in the intermediate nodes.

\subsection{Kolmogorov-Arnold Networks}
KANs are neural networks inspired by the Kolmogorov-Arnold representation theorem.
They address the constraints of the original design$-$which is limited to two layers and $2n+1$ nodes in the middle layer$-$ by stacking multiple KAN layers, each with an arbitrary number of nodes. Formally, a single KAN layer is defined as 
\begin{equation*}
    \Phi = \{\phi_{q,p}\}, \quad q = 1, \ 2, \ \cdots, \ n_{\text{in}}, \quad p = 1, \ 2, \ \cdots, \ n_{\text{out}},
\end{equation*} 
where the functions $\phi_{q, p}$ are univariate trainable activation functions, and \( n_{\text{in}} \) and \( n_{\text{out}} \) denote the dimensions of its inputs and outputs, respectively. Given  an $n_{in}$-dimensional input vector $\mathbf{x}=(x_1, \ x_2, \ \cdots, \ x_{n_{in}})$, the output of this KAN layer is defined by
\begin{equation*}
    \Phi(\mathbf{x}) = \left( \sum_{q=1}^{n_{in}} \phi_{q, 1} \left( x_q \right), \ \sum_{q=1}^{n_{in}} \phi_{q, 2} \left( x_q \right), \ \cdots, \ \sum_{q=1}^{n_{in}} \phi_{q, n_{out}} \left( x_q \right) \right) \in \mathbb{R}^{n_{out}}.
\end{equation*}
Then a generalized KAN with \( L \) layers is expressed as:
\begin{equation}
KAN(\mathbf{x}) = \left(\Phi^{(L-1)} \circ \cdots \circ \Phi^{(0)} \right)(\mathbf{x}),
\end{equation}
where each $\Phi^{(l)} := \{\phi_{q, p}^{(l)} \}$ represents the $l$-th KAN layer.
Notably, the structure derived from the Kolmogorov-Arnold representation theorem can be seen as a special case of a KAN with two layers and $2n+1$ nodes in the middle layer. By allowing arbitrary depths and widths, KANs substantially extend this framework’s flexibility.

{\textbf{Activation functions of KANs.}}
Conventional MLP-based neural networks typically employ a single activation function (e.g., ReLU, Sigmoid) shared across all nodes after a global linear transformation.
In contrast, KANs introduce residual activation functions that combine a smooth basis function $b(x)$ with a B-spline component.
Specifically, each node’s activation function in KAN is given by 
\begin{equation} \phi(x) = w_b b(x) + w_s \text{spline}(x),
\end{equation} 
where $w_b$ and $w_s$ are trainable scaling coefficients.
The basis function $b(x)$ is typically chosen to be the SiLU function,
\begin{equation*}
    b(x) = \frac{x}{1+e^{-x}},
\end{equation*}
and $\text{spline}(x)$ is a linear combination of B-spline basis functions given by
\begin{equation*}
    \text{spline}(x) = \sum_i c_i B_i (x),
\end{equation*}
where the coefficients $c_i$ are trainable parameters, and each B-spline basis function $B_i$ is of order $k$ with grid $G$.
In practice, these additional parameters $\{w_b, w_s, c_i\}$ are learned through standard backpropagation, providing robust flexibility while preserving interpretability.
Moreover, this activation function design facilitates locally precise approximations of complex relationships and enhances the model’s capacity for capturing highly nonlinear behavior.

\textbf{Training KANs.}
While the overall training procedure of KANs closely resembles that of MLP-based networks, it involves additional steps, such as node pruning and symbolification, to enhance interpretability. Specifically, we consider the VIX at time $t$, denoted by $V_t$, as our target variable, and seek to predict it from explanatory variables $x_{<t}$ observed prior to time $t$. In this setting, the predictive loss $\mathcal{L}_{\text{pred}}$ is defined as 
\begin{equation}
    \mathcal{L}_{\text{pred}} = \sum_t \left( V_t - KAN \left( x_{<t} \right) \right)^2,
\end{equation}
in the same manner to the mean-squared error used in conventional MLP-based neural network training.

According to the design of KANs, the number of nodes and layers can be arbitrarily large; however, in  many scientific and real-world tasks, most functions often have sparse compositional structures \cite{liu2024kan}. Motivated by the importance of interpretability in such contexts, KANs employ $L_1$-norm and entropy regularizations, both defined at the layer level, to facilitate sparse representations.
Specifically, the $L_1$-norm of a KAN layer $\Phi$ is defined by 
$\|\Phi\|_1 = \sum_{i,j} \|\phi_{ij}\|_1$ where \( \|\phi_{ij}\|_1 \) denotes the average magnitude of the activation function $\phi_{q, p}$ and is computed as $\| \phi_{q,p}\|_1 = \frac{1}{N_p} \sum_{s=1}^{N_p} \left| \phi_{q,p} (x^{(s)}) \right|$ for $N_p$ input data points $x^{(s)}$.
The entropy of \( \Phi \) is given by $S(\Phi) = -\sum_{q,p} \frac{\|\phi_{q,p}\|_1}{\|\Phi\|_1} \log\left(\frac{\|\phi_{q,p}\|_1}{\|\Phi\|_1}\right).$
Hence, the total training loss is
\begin{equation} \label{eqn: loss function}
\mathcal{L}_{\text{total}} = \mathcal{L}_{\text{pred}} + \lambda \left( \mu_1 \sum_{l=0}^{L-1} \left\|\Phi^{(l)} \right\|_1 + \mu_2 \sum_{l=0}^{L-1} S \left(\Phi^{(l)} \right) \right),
\end{equation}
where \( \lambda \) controls the overall regularization strength, and \( \mu_1, \mu_2 \) determine the weights for the \( L_1 \)-norm and entropy terms, respectively.

\begin{algorithm}[!t]
\caption{Training procedure of KANs}
\label{alg: kan training}
\KwIn{Explanatory variables $x_{<t}$ for forecasting $V_t$}
1. Train $KAN(x_{<t})$ to predict $V_t$ using the loss function \eqref{eqn: loss function}.

2. Prune low-importance edges and nodes.

3. Symbolify the trained B-spline-based activation functions, and incorporate affine transforms.

4. Fine-tune the affine parameters added in Step 3. 

\Return{The symbolic formula for $\hat{V_t}$}
\end{algorithm}

Once trained with these regularization terms, nodes whose incoming and outgoing values remain below a certain threshold are considered unimportant and are pruned. 
The pruned KAN then undergoes a symbolification step, in which B-spline-based activation functions are replaced by elementary (e.g., polynomial, exponential, sine) or domain-specific functions (e.g., Gaussian, Bessel) that closely approximate the learned activation functions.
At this step, one can visually identify the learned activation function, and a user with domain knowledge can manually replace them with a symbolic function.
Alternatively, a symbolic function can be automatically selected from a prespecified set of candidate functions.
Finally, an additional affine transformation is applied and these affine parameters are fine-tuned through further training. The overall training procedure is summarized in Algorithm \ref{alg: kan training}. For further details on the theoretical backgrounds and implementation, we refer to \cite{liu2024kan}.

\section{VIX Forecasting using KAN} \label{sec: KANs for vix forecasting}
\subsection{Data}

In this paper, we focus on forecasting the CBOE VIX using KANs. Since the VIX calculation method was revised in September 2003\footnote{The CBOE revised the calculation method for the VIX by switching its underlying asset from OEX to SPX and incorporating out-of-the-money options \cite{whaley2009understanding}.}, we employ the VIX data from January 2004 to December 2023 to ensure intertemporal consistency. Alongside the VIX data, we also utilize the S\&P 500 daily excess returns over the same period, using the 1-month U.S. Treasury bill yield obtained from the Federal Reserve Economic Data (FRED) as the risk-free rate. To align the scale with the VIX, the S\&P 500 daily excess returns are multiplied by 100.
% S\&P500 return: DataGuide

We construct three datasets  from historical VIX data to use as explanatory variables, each designed to examine whether particular characteristics of volatility can be observed. The first dataset uses five consecutive lagged values, \(\{V_{t-1}, V_{t-2}, V_{t-3}, V_{t-4}, V_{t-5}\}\), to predict the next VIX value \(V_t\). 
By focusing on the preceding five trading days, this dataset highlights short-term momentum.
The second dataset consists of non-uniform lags, \(\{V_{t-1}, V_{t-5}, V_{t-10}, V_{t-21}\}\), to investigate whether the short- (one day), medium- (one or two weeks), or long- (one month) range dependencies remain and provide predictability.
Finally, the third dataset comprises the previous VIX value, \(V_{t-1}\), with weekly average $V^{w}$, monthly average $V^{w}$, and quarterly average $V^{q}$, defined by
\begin{equation}\label{eqn: v_w v_m v_q}
V^{w}_{t-1} = \frac{1}{5}\sum_{i=1}^{5} V_{t-i}, 
\quad
V^{m}_{t-1} = \frac{1}{21}\sum_{i=1}^{21} V_{t-i},
\quad
V^{q}_{t-1} = \frac{1}{63}\sum_{i=1}^{63} V_{t-i},
\end{equation}
respectively.
We use this aggregated-statistics approach to determine whether the mean-reverting property of volatility across multiple horizons can be captured and, if so, to clarify the effects of short-, medium-, long-, and very long-term averages on volatility dynamics.
For convenience, we denote these datasets as \emph{Dataset 1}, \emph{Dataset 2}, and \emph{Dataset 3}, respectively, in the remainder of this paper.

To ensure the robustness of our experiments, we vary the proportions of training, validation, and test sets across three different configurations.
\emph{Period 1} allocates the data from 2004 to 2015 for training, 2016--2017 for validation, and 2018--2023 for testing, corresponding to a 6:1:3 ratio.
\emph{Period 2} applies a 7:1:2 ratio, and \emph{Period 3} uses 8:1:1.
Each of the three datasets is tested over all three periods, resulting in nine total experimental setups.
We denote a setup with \emph{Dataset i} and \emph{Period j} by \(D_iP_j\), where \(i \in \{1, \ 2, \ 3\}\) and \(j \in \{1, \ 2, \ 3\}\).

\subsection{Empirical Results}

\subsubsection{Experimental Setup}

% KAN hyperparameters
Before examining our empirical findings in detail, we describe the hyperparameters used for KAN in this study. All experiments, across every dataset and period, are conducted with the same hyperparameter settings.
Specifically, we use a two-layer KAN with two hidden nodes in the hidden layer. 
We adopt this parsimonious configuration because increasing the depth or the number of hidden nodes does not lead to any significant improvement or even any noticeable difference after the pruning and symbolification steps.
An analysis of the number of hidden nodes and layers is provided briefly in \ref{appendix: with deep depth and wide width}. 
The activation functions include B-spline components of order $k = 3$ with a grid size of 3.
For optimization, we employ the L-BFGS algorithm with a learning rate of 0.04. If the validation loss fails to improve for five consecutive epochs, we reduce the learning rate by a factor of 0.1; if it does not improve for ten consecutive epochs, we stop training activation functions and prune unimportant edges and nodes. We then replace the activation functions with symbolic functions that incorporate learnable affine parameters. Finally, these affine parameters are trained for 30 epochs with a learning rate of 0.0004. The regularization coefficient $\lambda$ for sparsification is set to zero during training, as unimportant edges and nodes are pruned effectively even without the help of this regularization.

% benchmark time-series models 
For comparison, we employ both statistical time-series models and MLP-based neural networks as benchmark methods.
The statistical models consist of a naive forward-filling method, ARMA and ARIMA approaches, and two variants of the HAR model.
The naive forward-filling method predicts $V_t$ simply by using the previous value (i.e., $\hat{V}_t = V_{t-1}$), motivated by the high persistence \cite{ding1993long} and clustering effects \cite{cont2001empirical} observed in volatility.
Despite its simplicity, this naive approach can show competitive performance in many scenarios.
The ARMA model combines AR and MA components to describe time-series behavior based on past values and errors, and the ARIMA model extends ARMA by applying differencing to model changes in the data rather than levels. 
To choose the AR parameter $p$, MA parameter $q$, and the order of differencing $d$, we conduct a grid search over $p, q \in \{0, 1, 2, 3, 4, 5\}$ and $d \in \{0, 1, 2\}$. The optimal values of $p$ and $q$ are selected based on the Akaike Information Criterion (AIC), and $d$ is determined using the KPSS test \cite{kwiatkowski1992testing}.
Through this process, we found that the ARMA(1,1) model and the ARIMA(1,1,1) model provides the optimal fit for our in-sample dataset.
We also compare with the HAR model, which utilizes averages of past volatility over multiple horizons.
Specifically, we employ two HAR models. One of these two model uses the previous VIX data, and weekly and monthly averages. 
This model is expected to capture the predictability induced by the short-, medium-, and long-term averages as in \cite{corsi2009simple}. 
The other model additionally includes a quarterly average to reflect the effect of very-long-term average.
These two HAR models are denoted by HAR(3) and HAR(4), respectively, and given by
\begin{equation}\label{eq:har_rv}
\hat{V}_{t} = c + \beta_1 V_{t-1} + \beta_2 V^{w}_{t-1} + \beta_3 V^{m}_{t-1} + \beta_4 V^{q}_{t-1},
\end{equation}
where $\beta_4 = 0$ in HAR(3), and $\beta_4\neq 0$ in HAR(4).
Note that, to predict $V_t$, the ARMA and ARIMA models employ \emph{Dataset 1}, and the HAR models use \emph{Dataset 3}. 
Thus, it is worthwhile to compare these models to KAN with \emph{Dataset 1} and \emph{Dataset 3}.

On the other hand, we also implement a MLP and a LSTM network as benchmarks. An MLP, which can be regarded as a nonlinear regressor \cite{goodfellow2016deep}, is an artificial neural network composed of input, hidden, and output layers. 
In our experiments, the MLPs are built with three hidden layers, each containing 100 units, where each linear transformation is followed by a LeakyReLU activation function. 
We train this model using the Adam optimizer with a learning rate of \(1 \times 10^{-3}\).
Meanwhile, the LSTM network, an extension of the RNN architecture, incorporates memory cells to learn long-term dependencies, which makes it more suitable for complex time-series forecasting. 
The LSTM models in our experiments consist of two stacked LSTM layers, each with 70 hidden units.
The final hidden states are passed through a two-layer fully connected network with a Sigmoid activation. We then scale its output by multiplying by 100.
This design allows the LSTM to handle intricate temporal features that may be difficult for simpler MLP models to capture.

Lastly, all the experiment in Section \ref{sec: KANs for vix forecasting} are performed using out-of-sample datasets.

\subsubsection{Analysis of KAN forecasting results} \label{sec: kan result analysis}

% We first concentrate on examining KAN forecasting results without yet comparing them to other benchmark models. The objectives here are twofold: to investigate both the results of the trained KANs and the effects of symbolification, and to statistically assess whether its forecast estimates exhibit bias or residual autocorrelation.

We begin by investigating the results of the trained KANs and the effects of symbolification without yet comparing them to other benchmark models.

\begin{figure}[p]
    \centering
    %---------------- Row j=1 (Train Period 1) ----------------%
    \begin{subfigure}[b]{0.3\textwidth}
        \centering
        \includegraphics[width=\linewidth]{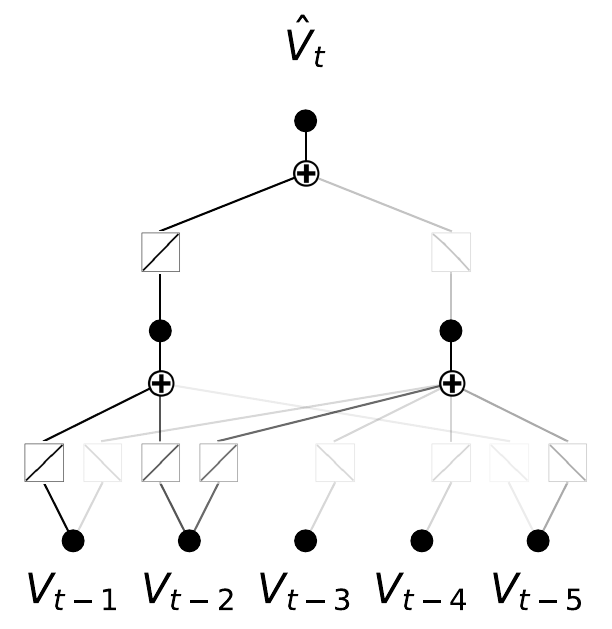}
        \caption{$D_1P_3$ before symbolification}
    \end{subfigure}
    \hfill
    \begin{subfigure}[b]{0.3\textwidth}
        \centering
        \includegraphics[width=\linewidth]{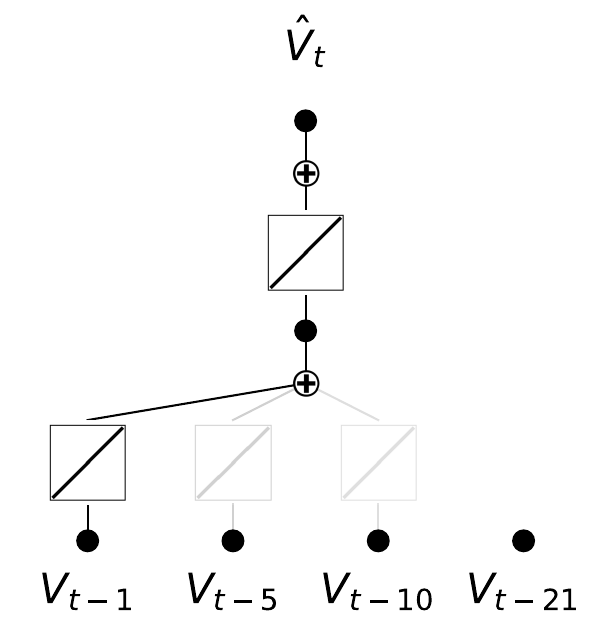}
        \caption{$D_2P_3$ before symbolification}
    \end{subfigure}
    \hfill
    \begin{subfigure}[b]{0.3\textwidth}
        \centering
        \includegraphics[width=\linewidth]{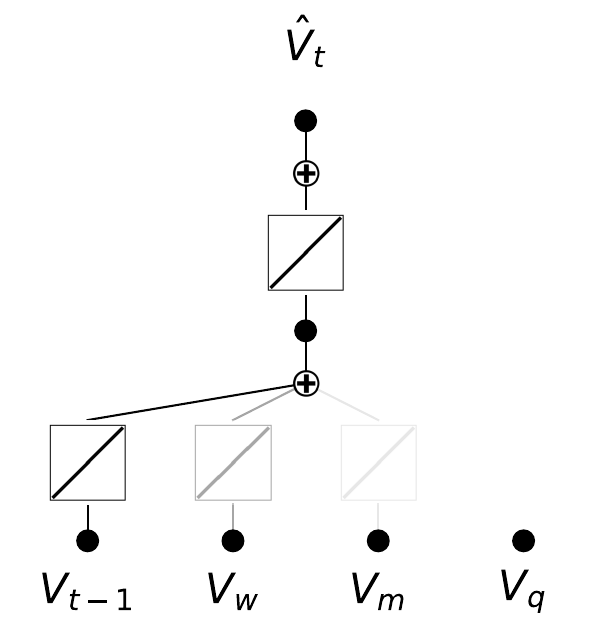}
        \caption{$D_3P_3$ before symbolification}
    \end{subfigure}
    
    \vskip\baselineskip

    %---------------- Row j=2 (Train Period 2) ----------------%
    \begin{subfigure}[b]{0.3\textwidth}
        \centering
        \includegraphics[width=\linewidth]{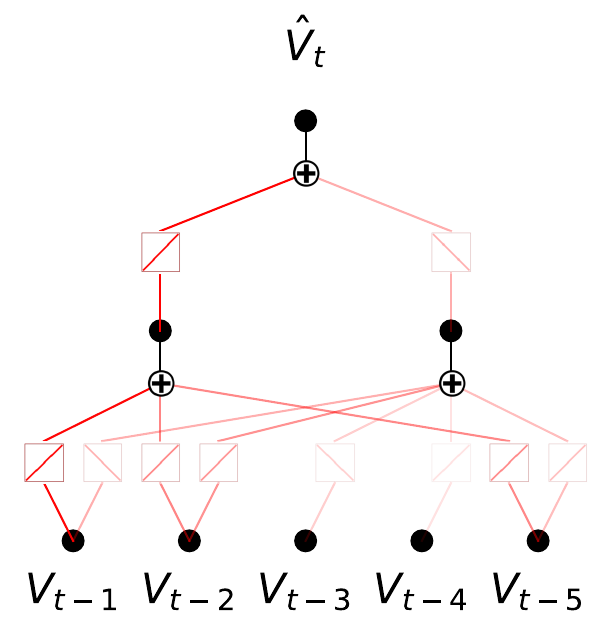}
        \caption{$D_1P_3$ after symbolification}
    \end{subfigure}
    \hfill
    \begin{subfigure}[b]{0.3\textwidth}
        \centering
        \includegraphics[width=\linewidth]{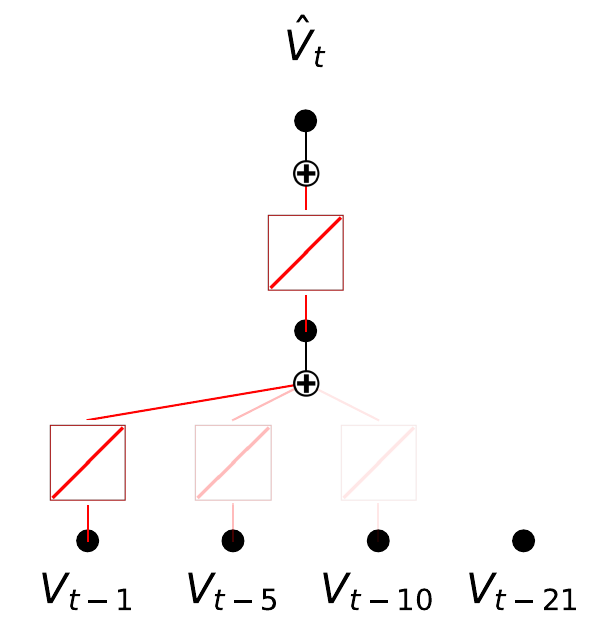}
        \caption{$D_2P_3$ after symbolification}
    \end{subfigure}
    \hfill
    \begin{subfigure}[b]{0.3\textwidth}
        \centering
        \includegraphics[width=\linewidth]{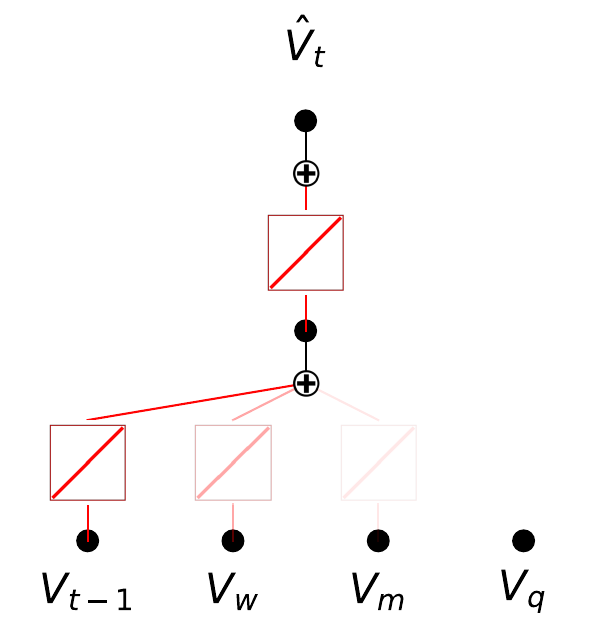}
        \caption{$D_3P_3$ after symbolification}
    \end{subfigure}
    \caption{KAN training results for \emph{Datasets 1--3} under \emph{Period 3} {before} (top row) and {after} (bottom row) symbolification. Black activation functions indicate the trained B-spline-based structure, whereas red activation functions depict the replaced symbolic functions. Vivid edges represent stronger effects, while fainter edges mean weaker ones.} \label{fig: KAN before after symbolification}
\end{figure}

Figure \ref{fig: KAN before after symbolification} illustrates the results of training KAN on three different datasets under \emph{Period 3}. The top row shows the learned activation functions before symbolification (i.e., the functions are still expressed as B-spline-based activation functions), while the bottom row illustrates the activation functions after symbolification (i.e., the functions have been replaced by symbolic functions).
From left to right, each column corresponds to \emph{Dataset 1}, \emph{Dataset 2}, and \emph{Dataset 3}. Since the outcomes for \emph{Periods 1} and \emph{2} are visually similar, they are presented in \ref{appendix: symbolification experiment P1 and P2} for brevity.

A notable observation from the top-row figures is that, in all cases including other periods in \ref{appendix: symbolification experiment P1 and P2}, KAN’s B-spline-based activation functions are trained to exhibit nearly linear shapes.
Recall that KAN’s activation functions (the black curves in the top-row figures) are designed to approximate arbitrary univariate continuous functions via splines. 
However, under the configurations and explanatory variables used in this study, linear forms consistently emerge. 
This suggests that, for predicting the VIX, every explanatory variable we tested primarily exerts \emph{linear} influence on forecast performance.
It also aligns with prior research indicating that simpler and more parsimonious models often suffice to capture essential characteristics of noisy financial time-series data \cite{diebold1998elements, box2015time, hamilton2020time}.
Moverover, this observation indirectly supports the linearity assumptions commonly made by statistical time-series models such as ARMA, ARIMA, and HAR, while simultaneously  highlighting KAN's distinctive capability.
Whereas these statistical models presuppose linearity, KAN identifies such structures directly from data, imposing no prior assumptions on its functional form.
This advantage allows researchers to uncover data-driven insights.

Another important observation is the automatic pruning of less influential edges and nodes. In these figures, edges are drawn more vividly when their impact is high and faintly if their impact is low. 
Nodes and edges with importance below a certain threshold are pruned and removed from the final network. Notably, in \emph{Dataset 2} and \emph{Dataset 3}, the network prunes unimportant nodes in the middle layer so that only a single influential node remains; similarly, for $V_{t-21}$ in \emph{Dataset 2} and $V_q$ in \emph{Dataset 3}, the pruned edges starting from these nodes make them disconnected from the prediction results. This automatic selection of meaningful edges and nodes highlights KAN’s capacity to return a sparse and parsimonious representation of the data.

Finally, based on the visual identification of these activation functions, we specify a set of candidate symbolic functions: $y=x$ and $y=0$. During symbolification, the trained activation functions are replaced by the closest symbolic function, followed by the incorporation of affine transformations. These affine parameters are fine-tuned afterward. The bottom-row figures (in red) depict these final activation functions, demonstrating how symbolification simplifies the learned representation.

We next assess the effects of symbolification on forecasting performance, as given in Table \ref{tab:comparison_r2}. The table compares the $R^2$ values of KAN forecasts before and after symbolification for each \emph{Dataset} and \emph{Period}.
Notably, for all cases except $D_2P_3$, the $R^2$ value improves slightly once the B-spline-based activation functions are replaced by simpler symbolic functions. 
This finding reinforces the notion that a more parsimonious model can work effectively for financial time-series data that exhibit a low signal-to-noise ratio.
In other words, activation functions may have been subtly overfitted and deviating from a linear shape during training, but symbolification corrects these deviations and enhances predictability in most cases. 

We also note that \emph{Dataset 3} consistently yields the highest $R^2$ across all three periods, with \emph{Dataset 1} a close second and \emph{Dataset 2} somewhat lower.
This result suggests that recent VIX values and weekly, monthly, and quarterly averages (\emph{Dataset 3}) provide more stable and robust predictive information than non-consecutive lagged values (\emph{Dataset 2}). 
By aggregating VIX data over multiple horizons, \emph{Dataset 3} partially averages out daily noise, making its signal clearer and capturing longer-range dynamics more reliably.
We further confirm these findings with additional benchmark models and performance metrics in Section \ref{sec: comparison with benchmarks}.

\begin{table}[!tb]
\centering
\caption{Comparison of \(R^2\) of KAN results before and after symbolification across different \emph{Dataset}s and \emph{Period}s.}
\label{tab:comparison_r2}
\begin{tabular}{c c c c}  
\toprule
\textbf{Dataset} & \textbf{Period} & \textbf{KAN before symbolification} & \textbf{KAN after symbolification} \\
\midrule

% --------------------------
% Dataset = 1
% --------------------------
\multirow{3}{*}{1} 
  & 1 & 0.9302 & \textbf{0.9306} \\ 
  & 2 & 0.9285 & \textbf{0.9286} \\ 
  & 3 & 0.9398 & \textbf{0.9400} \\ 
\midrule

% --------------------------
% Dataset = 2
% --------------------------
\multirow{3}{*}{2} 
  & 1 & 0.9287 & \textbf{0.9291} \\
  & 2 & 0.9262 & \textbf{0.9264} \\ 
  & 3 & \textbf{0.9397} & 0.9392 \\ 
\midrule

% --------------------------
% Dataset = 3
% --------------------------
\multirow{3}{*}{3} 
  & 1 & 0.9303 & \textbf{0.9307} \\ 
  & 2 & 0.9285 & \textbf{0.9286} \\ 
  & 3 & 0.9403 & \textbf{0.9403} \\ 

\bottomrule
\end{tabular}
\end{table}

The most significant aspect of symbolification is that it provides a closed-form expression composed of symbolic functions, thereby enabling interpretable forecasting.
Table \ref{tab:vix_equations} presents the closed-form symbolic expressions of $\hat{V}_t$ across different \emph{Dataset}s and \emph{Period}s, which correspond to the bottom row plots in Figure \ref{fig: KAN before after symbolification}.
In each case, the B-spline-based activation functions are replaced by simpler functional forms, and the corresponding coefficients for each explanatory variable are given explicitly.

\begin{table}[p]
\centering
\caption{Closed-form expressions \(\hat{V}_t\) across different \emph{Dataset}s and \emph{Period}s}
\label{tab:vix_equations}

\begin{tabular}{c c l}  
\toprule
\textbf{Dataset} & \textbf{Period} & \multicolumn{1}{c}{\textbf{Closed-form expression}} \\ 
\midrule

% --------------------------
% Dataset = 1
% --------------------------
\multirow{3}{*}{1} 
  & 1 & 
  \( 
    \hat{V}_t 
    = 0.8482 \cdot V_{t-1}
    + 0.0536 \cdot V_{t-2}
    + 0.0421 \cdot V_{t-3}
    - 0.0447 \cdot V_{t-4}
    + 0.0816 \cdot V_{t-5}
    + 0.3779
  \)\\
  \cmidrule(lr){2-3}
  & 2 &
  \(
    \hat{V}_t 
    = 0.8785 \cdot V_{t-1}
    + 0.0160 \cdot V_{t-2}
    + 0.0561 \cdot V_{t-3}
    - 0.0321 \cdot V_{t-4}
    + 0.0865 \cdot V_{t-5}
    + 0.2345
  \)\\
  \cmidrule(lr){2-3}
  & 3 & 
  \(
    \hat{V}_t 
    = 0.8584 \cdot V_{t-1}
    + 0.0206 \cdot V_{t-2}
    + 0.0632 \cdot V_{t-3}
    - 0.0374 \cdot V_{t-4}
    + 0.0801 \cdot V_{t-5}
    + 0.2793
  \)\\

\midrule

% --------------------------
% Dataset = 2
% --------------------------
\multirow{3}{*}{2} 
  & 1 & 
  \(
    \hat{V}_t 
    = 0.8661 \cdot V_{t-1}
    + 0.0634 \cdot V_{t-5}
    + 0.0655 \cdot V_{t-10}
    + 0.0712
  \)\\
  \cmidrule(lr){2-3}
  & 2 & 
  \(
    \hat{V}_t 
    = 0.8800 \cdot V_{t-1}
    + 0.0733 \cdot V_{t-5}
    + 0.0359 \cdot V_{t-10}
    + 0.1950
  \)\\
  \cmidrule(lr){2-3}
  & 3 & 
  \(
    \hat{V}_t 
    = 0.8840 \cdot V_{t-1}
    + 0.0742 \cdot V_{t-5}
    + 0.0344 \cdot V_{t-10}
    + 0.1099
  \)\\

\midrule

% --------------------------
% Dataset = 3
% --------------------------
\multirow{3}{*}{3} 
  & 1 & 
  \(
    \hat{V}_t 
    = 0.8297 \cdot V_{t-1}
    + 0.1477 \cdot V_{\text{w}}
    + 0.4756
  \)\\
  \cmidrule(lr){2-3}
  & 2 & 
  \(
    \hat{V}_t 
    = 0.8189 \cdot V_{t-1}
    + 0.1660 \cdot V_{\text{w}}
    + 0.2749
  \)\\
  \cmidrule(lr){2-3}
  & 3 & 
  \(
    \hat{V}_t 
    = 0.8255 \cdot V_{t-1}
    + 0.1292 \cdot V_{\text{w}}
    + 0.0377 \cdot V_{\text{m}}
    + 0.1025
  \)\\

\bottomrule
\end{tabular}
\end{table}

A consistent pattern emerges across all cases: $V_{t-1}$ exerts a dominant influence, typically with a coefficient exceeding 0.8.
This overall observation indicates that KAN successfully captures the well-known high persistence and clustering effect of the VIX. 
Turning specifically to \emph{Dataset 1}, the coefficient of $V_{t-5}$ consistently shows the second-largest magnitude, whereas $V_{t-2}$, $V_{t-3}$, and $V_{t-4}$ appear less consistent and persistent. 
Nonetheless, the coefficients of these three intermediate variables retain the same signs, and only $V_{t-4}$ affects the prediction negatively.
In \emph{Dataset 2}, the term $V_{t-21}$ is pruned in every period.
Meanwhile, in \emph{Dataset 3}, the very-long-term average $V_q$ is pruned in all \emph{Period}s, and even the long-term average $V_m$ is pruned in certain periods.
These observations suggest that long- or very-long-range dependence may contribute less to forecasting the VIX than short- and medium-range variables.

It is also noteworthy that in \emph{Dataset 3}, the contribution of $V_{t-1}$ may appear smaller than in the other datasets. However, this is because $V_{t-1}$ is partially embedded in the weekly or monthly averages. 
For instance, in \emph{Period 1} of \emph{Dataset 3}, the coefficient of the weekly average $V_w$ is estimated at 0.1477. 
Thus, the total contribution from $V_{t-1}$ is calculated as $0.8290 + (0.1472 / 5) = 0.8584$, which is comparable with the results of \emph{Dataset 1}.
Additionally, the fact that the coefficient of $V_w$ in \emph{Dataset 3} consistently exceeds 0.1 and surpasses all other variables except $V_{t-1}$ demonstrates that aggregated averages provide significant predictability.

Lastly, we empirically demonstrate that the VIX can be modeled with mean-reverting properties using the closed-form expression derived from \emph{Dataset 3}. Let us first examine \emph{Period 1} with \emph{Dataset 3}. Here, we denote the difference between the predicted and previous values, $\hat{V}_t - V_{t-1}$, as $\Delta V_t$ which is given by
\begin{equation} \label{eqn: delta Vt period 1}
\begin{aligned}
\Delta V_t %&= \hat{V}_t - V_{t-1} \\
&= \left( 0.8290 V_{t-1} + 0.1472 V_w + 0.4866 \right) - V_{t-1} \\
&=  0.1472 \left(V_w - V_{t-1} \right) + \left(- 0.0238 V_{t-1} + 0.4866 \right)% := 0.1472 (V_w - V_{t-1}) + \epsilon_t.
\end{aligned}
\end{equation}
% \begin{equation} \label{eqn: delta Vt period 1}
% \begin{aligned}
% \hat{V}_t &= 0.8290 V_{t-1} + 0.1472 V_w + 0.4866 \\
% &= V_{t-1} + 0.1472 \left(V_w - V_{t-1} \right) + \left(- 0.0238 V_{t-1} + 0.4866 \right)% := 0.1472 (V_w - V_{t-1}) + \epsilon_t.
% \end{aligned}
% \end{equation}
The historical average of the VIX during the training period of \emph{Period 1} (2004–2015) is estimated to be 20.34, and the mean of the last term $(- 0.0238 V_{t-1} + 0.4866)$ is calculated to be 0.0025.
Similarly, for \emph{Period 2} and \emph{Period 3}, the mean values of the corresponding terms in the equations for $\Delta V_t$ are calculated as -0.0164 and -0.0177, respectively, indicating that the the value of these terms is nearly zero. 
Based on our empirical observations, we can 
propose the following model of $\Delta V_t$ capturing the mean-reverting property of the VIX as
\begin{equation} \label{eqn: delta Vt}
    \Delta V_t %= \hat{V}_t - V_{t-1} 
    = \kappa (\theta - V_{t-1}) + \epsilon_t,
\end{equation}
% \begin{equation} \label{eqn: delta Vt}
%     \hat{V}_t = V_{t-1} + \kappa (\theta - V_{t-1}) + \epsilon_t,
% \end{equation}
where $\kappa$ and $\theta$ are constants. Here, $\epsilon_t$ may or may not depend on $V_{t-1}$, and is assumed to have zero-mean.
A stochastic process whose dynamics is given as \eqref{eqn: delta Vt} is known to have a mean-reverting property, and $\kappa$ and $\theta$ represent the mean-reverting speed and the mean-level, respectively (e.g., see \cite{heston1993closed,scott1987option}).

If we let $(- 0.0238 V_{t-1} + 0.4866)$ in \emph{Period 1} be $\epsilon_t$, then the equation \eqref{eqn: delta Vt period 1} becomes identical to \eqref{eqn: delta Vt}.
Consequently, the closed-form expressions derived by KAN with \emph{Dataset 3} can be interpreted as representing a typical mean-reverting process reverting to its medium- or long-term average.
Note that this interpretation arises from KAN's ability to provide symbolic closed-form expressions, while such interpretability is generally unattainable in black-box MLP-based models.

\subsubsection{Statistical assessment of KAN as a forecasting model} \label{sec: statistical test for kan}

To examine the statistical properties and significance of our fitted prediction model, we employ two tests commonly used in financial time-series forecasting$-$the Mincer--Zarnowitz test \cite{mincer1969evaluation} and the Durbin--Watson test \cite{durbin1992testing1, durbin1992testing2}$-$ and present the results in Table~\ref{tab:mz_dw_results}. The Mincer--Zarnowitz approach examines whether a forecast is unbiased by regressing
\begin{equation} \label{eqn: MZ equation}
    V_t = \alpha + \beta \hat{V_t} + u_t,
\end{equation}
for residual $u_t$, and constant $\alpha$ and $\beta$, and testing whether $\alpha=0$ and $\beta=1$ \cite{mincer1969evaluation}. In our setting, the F-statistics remain well below the usual critical thresholds, and $p$-values consistently exceed conventional significance levels (e.g., 0.10), indicating insufficient evidence to reject the null hypothesis $(\alpha, \beta) = (0, 1)$. In other words, there is no strong indication that $\alpha \neq 0$ or $\beta \neq 1$. 
Meanwhile, the Durbin--Watson statistic ranges between approximately 1.799 and 2.180 for all datasets and forecast periods.
This statistic typically falls between 0 and 4, with values near 2 indicating no first-order autocorrelation.
Empirical studies often use 1.5--2.5 as a practical threshold range for concluding that no substantial first-order serial correlation remains.
In our results, the statistic falls comfortably within that interval, suggesting no problematic first-order autocorrelation in the residuals. Overall, these results suggest that the model’s forecasts do not deviate significantly from unbiasedness, nor do the residuals exhibit problematic autocorrelation.

\begin{table}[!t]
    \centering
    \caption{
      Results of the Mincer--Zarnowitz test and Durbin--Watson test are presented.
      The Mincer--Zarnowitz columns display 
      the $F$-statistic and its $p$-value under the null hypothesis $(\alpha,\beta)=(0,1)$. The      Durbin--Watson statistic within $[1.5,\,2.5]$ typically indicates 
      no significant first-order residual autocorrelation.
    }
    \label{tab:mz_dw_results}
    \begin{tabular}{ccccc}
    \toprule
    \multirow{2}{*}{\textbf{Dataset}} 
      & \multirow{2}{*}{\textbf{Period}} 
      & \multicolumn{2}{c}{\textbf{Mincer--Zarnowitz}} 
      & \multirow{2}{*}{\textbf{Durbin--Watson statistic}} \\
    \cmidrule(lr){3-4}
     &  & \textbf{$F$-statistic} & \textbf{$p$-value} &  \\
    \midrule
    \multirow{3}{*}{1} & 1 & 1.9642 & 0.1406 & 2.098 \\
     & 2 & 1.6492 & 0.1927 & 2.178 \\
     & 3 & 0.1914 & 0.8258 & 1.799 \\
    \midrule
    \multirow{3}{*}{2} & 1 & 1.7725 & 0.1703 & 2.167 \\
     & 2 & 1.5861 & 0.2052 & 2.157 \\
     & 3 & 1.1493 & 0.3177 & 1.824 \\
    \midrule
    \multirow{3}{*}{3} & 1 & 0.1938 & 0.8238 & 2.156 \\
     & 2 & 1.9326 & 0.1453 & 2.180 \\
     & 3 & 0.1279 & 0.8800 & 1.806 \\
    \bottomrule
    \end{tabular}
\end{table}

\subsubsection{Capturing the leverage effect}

Having established that KAN effectively replicates volatility clustering and mean-reverting behavior, we next examine its capacity to capture the leverage effect. 
In particular, we aim to determine whether excess returns of the underlying asset (S\&P 500) help explain remaining variability in VIX forecasts once past VIX data have been accounted for.
To this end, we consider the augmented model:
\begin{equation} \label{eqn: vt = kan(hat v, R e)}
    {V}_t = \tilde{V}_t + e_t, \quad \text{where} \quad \tilde{V}_t := KAN(\hat{V}_t, R_{t-1}^e).
\end{equation}
Here, $\hat{V}_t$ is the KAN-based forecast derived exclusively from past VIX observations using \emph{Datasets 1--3}, $R_{t-1}^e$ is the excess return of the underlying asset at $t-1$, and $e_t$ is a residual term associated with $\tilde{V}_t$.

Under the hypothesis that the KAN models in Section \ref{sec: kan result analysis} are trained optimally on their given explanatory variables (i.e., $\hat{V}_t$ represents the best possible prediction from historical VIX alone), we attempt to explain the remaining variability stemming from the stock market via $R_{t-1}^e$.
If this hypothesis holds, then the KAN model in \eqref{eqn: vt = kan(hat v, R e)} would return $\hat{V}_t$ unchanged and the effect of the excess return would be expressed through a suitable function during training. 
In other words, the VIX may be written as
\begin{equation} \label{eqn: vt decomposition with leverage effect}
    V_t = \tilde{V}_t + e_t = \left(\hat{V}_t + \psi (R_{t-1}^e) \right) + e_t,
\end{equation}
where $\psi$ is a function KAN learns for excess returns, and $e_t$ remains the irreducible noise.
From the perspective of the Mincer-Zarnowitz test results in Section \ref{sec: statistical test for kan}, which did not reject the null hypothesis $\alpha=0$ and $\beta=1$ in \eqref{eqn: MZ equation}, this expression suggests that the error term $u_t$ in \eqref{eqn: MZ equation} can be decomposed into $\psi(R_{t-1}^e)$ and the irreducible error $e_t$.

\begin{figure}[!tb] 
    \centering
    %---------------- Row j=1 (Train Period 1) ----------------%
    \begin{subfigure}[b]{0.25\textwidth}
        \centering
        \includegraphics[width=\linewidth]{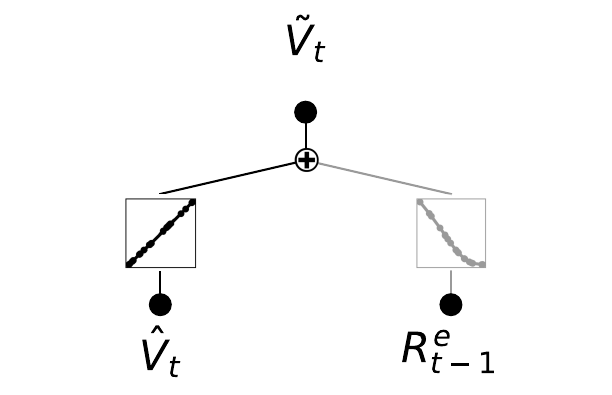}
        \caption{$D_1P_1$}
    \end{subfigure}
\hspace{0.01\textwidth}
    \begin{subfigure}[b]{0.25\textwidth}
        \centering
        \includegraphics[width=\linewidth]{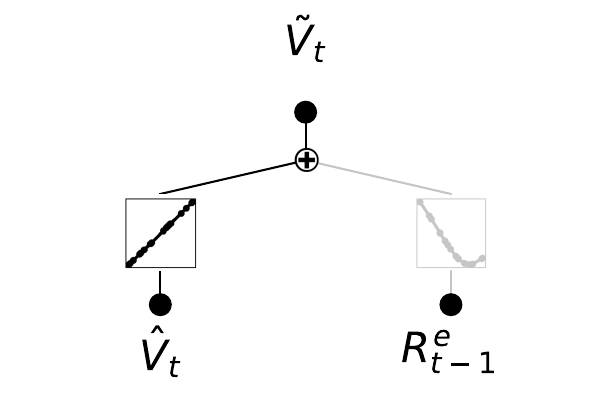}
        \caption{$D_1P_2$}
    \end{subfigure}
\hspace{0.01\textwidth}
    \begin{subfigure}[b]{0.25\textwidth}
        \centering
        \includegraphics[width=\linewidth]{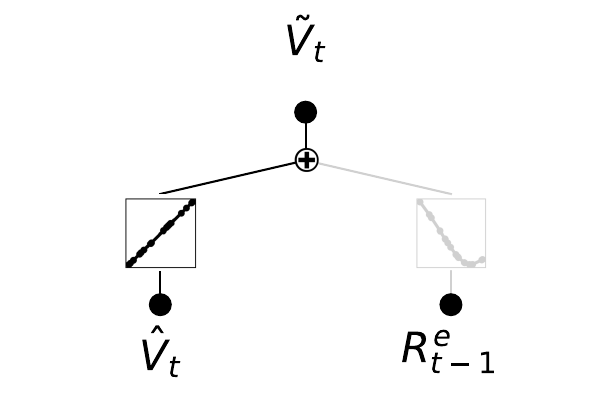}
        \caption{$D_1P_3$}
    \end{subfigure} \\
    \begin{subfigure}[b]{0.25\textwidth}
        \centering
        \includegraphics[width=\linewidth]{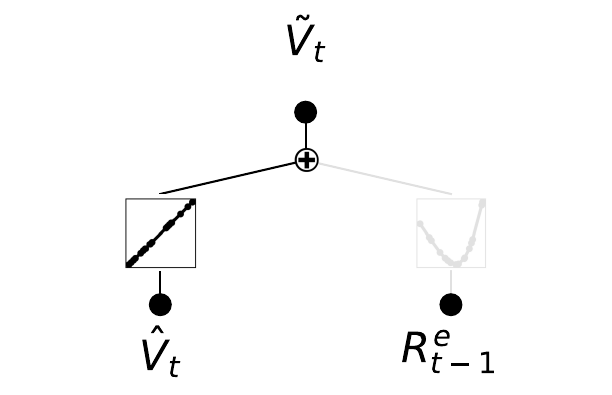}
        \caption{$D_2P_1$}
    \end{subfigure}
\hspace{0.01\textwidth}
    \begin{subfigure}[b]{0.25\textwidth}
        \centering
        \includegraphics[width=\linewidth]{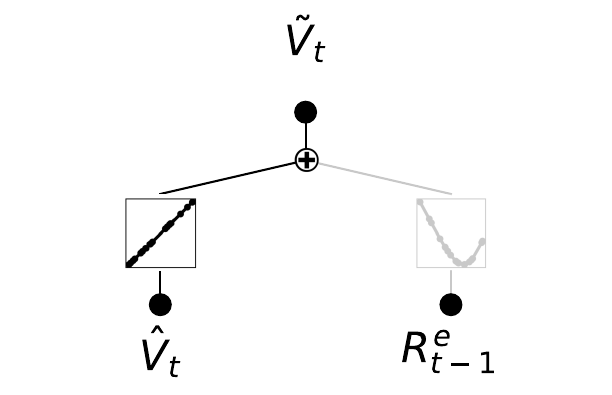}
        \caption{$D_2P_2$}
    \end{subfigure}
\hspace{0.01\textwidth}
    \begin{subfigure}[b]{0.25\textwidth}
        \centering
        \includegraphics[width=\linewidth]{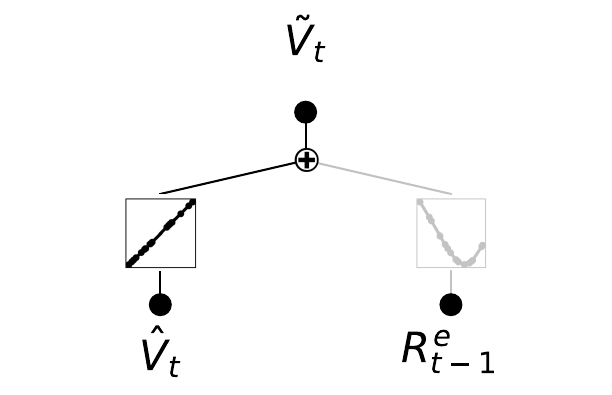}
        \caption{$D_2P_3$}
    \end{subfigure} \\
    \begin{subfigure}[b]{0.25\textwidth}
        \centering
        \includegraphics[width=\linewidth]{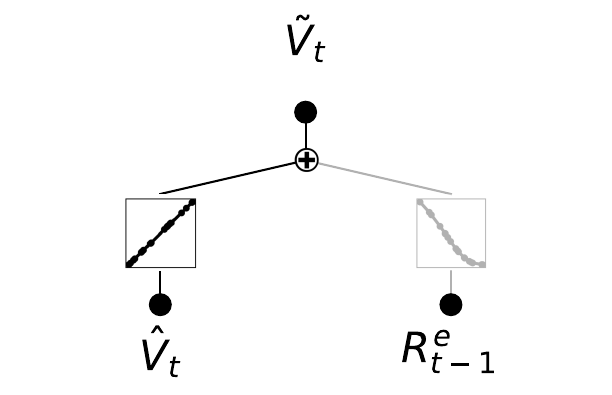}
        \caption{$D_3P_1$}
    \end{subfigure}
\hspace{0.01\textwidth}
    \begin{subfigure}[b]{0.25\textwidth}
        \centering
        \includegraphics[width=\linewidth]{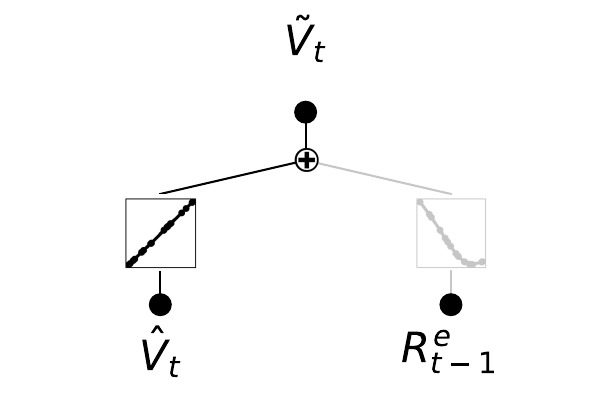}
        \caption{$D_3P_2$}
    \end{subfigure}
\hspace{0.01\textwidth}
    \begin{subfigure}[b]{0.25\textwidth}
        \centering
        \includegraphics[width=\linewidth]{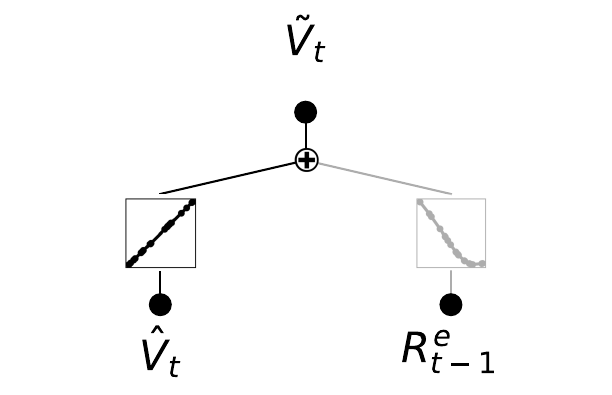}
        \caption{$D_3P_3$}
    \end{subfigure} 
    \caption{Training results of $KAN(\hat{V}_t, R_{t-1}^e)$ across \emph{Dataset}s and \emph{Period}s. To clearly see the activation functions, the contrast of the figures is increased by three times compared to Figure \ref{fig: KAN before after symbolification}.
}
    \label{fig: kan leverage effect} 
\end{figure}

We design a straightforward KAN to explore how $R_{t-1}^e$ affects the remaining forecast error. Specifically, we feed both $\hat{V}_t$ and $R_{t-1}^e$ into a one-layer KAN, where each passes through its own B-spline-based activation function to construct $\tilde{V}_t$.
Figure \ref{fig: kan leverage effect} illustrates the learned activation functions for all three \emph{Dataset}s across three \emph{Period}s.
Notably, $\hat{V}_t$ consistently exhibits an almost-linear shape whereas the activation functions for $R_{t-1}^e$ tend to decrease in every case except for \emph{Dataset 2} in \emph{Period 1}.
This downward slope is a hallmark of the leverage effect as lower excess returns correspond to a higher predicted VIX.
Occasionally, for extremely large positive returns, the functions flatten or even curve upward.
This behavior implies that the model predicts high volatility following such extreme returns, which seems a reasonable outcome given the typical market reactions to drastic price movements.

\begin{table}[!b]
    \centering
    \caption{\(R^2\) values of $\tilde{V}_t$ and their improvement over $\hat{V}_t$ in parenthesis, and closed-form expressions after symbolification. } 
    \label{tab:comparison_r2_leverage_effect}
\begin{tabular}{c c c c}  
\toprule
\textbf{Dataset} & \textbf{Period} & $\tilde{V}_t$ $(\tilde{V}_t - \hat{V}_t)$ & \textbf{Closed-form expression 
 after symbolification} \\
\midrule

% --------------------------
% Dataset = 1
% --------------------------
\multirow{3}{*}{1} 
  & 1 & 0.9314 (0.0007) & $\tilde{V}_t = 0.9974 \cdot \hat{V}_t - 0.0662 \cdot R^e_{t-1} - 0.0393$ \\ 
  & 2 & 0.9297 (0.0011) & $\tilde{V}_t = 1.0017 \cdot \hat{V}_t - 0.0339 \cdot R^e_{t-1} - 0.0977$ \\  
  & 3 & 0.9401 (0.0002) & $\tilde{V}_t = 1.0035 \cdot \hat{V}_t - 0.0284 \cdot R^e_{t-1} - 0.0870$ \\ \midrule

% --------------------------
% Dataset = 2
% --------------------------
\multirow{3}{*}{2} 
  & 1 & 0.9301 (0.0010) & $\tilde{V}_t = 1.0041 \cdot \hat{V}_t + 0.0043 \cdot R^e_{t-1} - 0.0982$ \\ 
  & 2 & 0.9281 (0.0017) & $\tilde{V}_t = 1.0026 \cdot \hat{V}_t - 0.0237 \cdot R^e_{t-1} - 0.0737$ \\  
  & 3 & 0.9396 (0.0004) & $\tilde{V}_t = 0.9885 \cdot \hat{V}_t - 0.0294 \cdot R^e_{t-1} + 0.2487$ \\ \midrule

% --------------------------
% Dataset = 3
% --------------------------
\multirow{3}{*}{3} 
  & 1 & 0.9318 (0.0011) & $\tilde{V}_t = 0.9921 \cdot \hat{V}_t - 0.0505 \cdot R^e_{t-1} + 0.0665$ \\ 
  & 2 & 0.9295 (0.0009) & $\tilde{V}_t = 1.0041 \cdot \hat{V}_t - 0.0355 \cdot R^e_{t-1} - 0.1278$ \\  
  & 3 & 0.9407 (0.0004) & $\tilde{V}_t = 1.0006 \cdot \hat{V}_t - 0.0523 \cdot R^e_{t-1} - 0.0212$ \\
\bottomrule
\end{tabular}
\end{table}

Table \ref{tab:comparison_r2_leverage_effect} reports the $R^2$ values for $\tilde{V}_t$ and their improvements over $\hat{V}_t$. 
In every case, incorporating $R_{t-1}^e$ robustly enhances forecasting accuracy across all \emph{Dataset}s and \emph{Period}s.
These improvements suggest that latent information in excess return data helps KAN to refine VIX estimates.
To further quantify the contribution of excess returns, we replace the activation functions for $R_{t-1}^e$ with a simple linear function $y=-x$.
The resulting closed-form expressions are presented in Table \ref{tab:comparison_r2_leverage_effect}.
Although we acknowledge that the trained activation functions for $R_{t-1}^e$ cannot be strictly regarded as $y=-x$ and that this replacement may introduce errors, our objective is to provide a quantitative assessment of how $R_{t-1}^e$ influences $\tilde{V}_t$.
As shown in the table, the coefficient of $\hat{V}_t$ remains close to one in all cases, aligning with our theoretical expectations in \eqref{eqn: vt decomposition with leverage effect}.
Additionally, the coefficients for $R_{t-1}^e$ range from $-0.02$ to $-0.07$ in most cases except for \emph{Dataset 2} in \emph{Period 1}. 
This result suggests a nonnegligible negative impact of $R_{t-1}^e$ on forecasting the VIX, and further demonstrates KAN's ability to capture the leverage effect effectively.

The anomalous positive coefficient of $R_{t-1}^e$ for \emph{Dataset 2} in \emph{Period 1} is attributed to the corresponding activation function before symbolification (see Figure \ref{fig: kan leverage effect}). 
In the figure, the KAN is trained to predict substantially higher $V_t$ values for large positive $R_{t-1}^e$.
As a result, the learned activation function becomes nonlinear and does not align with the linear $y=-x$, leading to a less meaningful outcome upon symbolification.
% Although the trained activation functions for $R_{t-1}^e$ predominantly exhibit a downward slope, symbolifying them with linear functions may not be inappropriate.
Moreover, automatically symbolifying them using the closest symbolic function can undermine interpretability\footnote{In some trials, the activation functions were replaced by a sine function or error function, yet these replacements seem meaningless in practice.}.
For this reason, we exclude these adjusted models from forecasting performance comparisons with other benchmarks, as interpretability remains a primary motivation for adopting KAN in this study.

\subsubsection{Comparison with traditional and deep learning benchmark methods} \label{sec: comparison with benchmarks}

We compare forecasting performances of KAN and various benchmark methods: traditional statistical time-series models (forward filling, ARMA, ARIMA, HAR), and neural network models (MLP, LSTM).
For notational convenience, the neural network model (KAN, MLP, LSTM) trained on \emph{Dataset i} is denoted as \emph{model-$D_i$} for $i=1,2,3$.
To ensure a robust and comprehensive assessment of model performance, we employ several metrics: mean squared error (MSE), mean absolute error (MAE), mean absolute percentage error (MAPE), \(R^2\), and quasi-likelihood (QLIKE).
Their definitions are given by
\begin{equation*}
\begin{gathered}
\text{MSE} = \frac{1}{N_{test}} \sum_{t=1}^{N_{test}} \left( V_t - \hat{V}_t \right)^2, \quad
\text{MAE} = \frac{1}{N_{test}} \sum_{t=1}^{N_{test}} \left| V_t - \hat{V}_t \right|, \quad
\text{MAPE} = \frac{1}{N_{test}} \sum_{t=1}^{N_{test}} \left| \frac{V_t - \hat{V}_t}{V_t} \right| \times 100, \\
R^2 = 1 - \frac{\sum_{t=1}^{N_{test}} \left( V_t - \hat{V}_t \right)^2}{\sum_{t=1}^{N_{test}} \left( V_t - \bar{V} \right)^2}, \qquad
\text{QLIKE} = \frac{1}{N_{test}} \sum_{t=1}^{N_{test}} \left[ \frac{V_t}{\hat{V}_t} - \ln \frac{V_t}{\hat{V}_t} - 1 \right],
\end{gathered}
\end{equation*}
where \(N_{test}\) is the number of test data points, \(\bar{V}\) denotes the mean of the VIX index in the test period, and \(V_t\) and \(\hat{V}_t\) are the actual and forecasted VIX values at time \(t\), respectively.

\begin{table}[!b]
\centering
\caption{Forecasting accuracy comparison of various models (\emph{Period 1}). \textbf{Bold} and \underline{underline} values indicate the best and second-best performances, respectively.}
\label{tab:performance_comparison_p1}
\setlength{\tabcolsep}{5pt}
\renewcommand{\arraystretch}{1.2}
\begin{tabular}{lrrrrrr}
\toprule
\textbf{Model} & \textbf{MSE} & \textbf{MAE} & \textbf{MAPE} & $\bm{R^2}$ & \textbf{QLIKE} & \textbf{\# params} \\
\midrule
Forward filling & 4.5302 & 1.2425 & 5.5466 & 0.9285 & \underline{0.0143} & - \\
ARMA(1, 1)      & 6.9365 & 1.6635 & 7.6243 & 0.8906 & 0.5433 & 4 \\
ARIMA(1, 1, 1)  & 7.1332 & 1.6745 & 7.6344 & 0.8875 & 0.5465 & 4 \\
HAR(3)       & {4.4337} & {1.2228} & {5.4967} & {0.9301} & 0.0144 & 4 \\
HAR(4)       & 4.4347 & 1.2229 & 5.4970 & 0.9300 & 0.0144 & 5 \\
\midrule
MLP-$D_1$  & 4.4262 & 1.2317 & 5.5294 & 0.9302 & 0.0154 & 20901 \\
MLP-$D_2$          & 4.6329 & \underline{1.2180} & \underline{5.4044} & 0.9269 & 0.0151 & 20801 \\
MLP-$D_3$           & 4.7031 & \textbf{1.2116} & \textbf{5.3792} & 0.9258 & 0.0159 & 20801 \\
LSTM-$D_1$      & 4.8150 & 1.2396 & 5.5124 & 0.9240 & 0.0163 & 63841 \\
LSTM-$D_2$         & 4.8059 & 1.2908 & 5.7796 & 0.9242 & 0.0156 & 63561 \\
LSTM-$D_3$          & 4.9712 & 1.2552 & 5.5850 & 0.9216 & 0.0157 & 63561 \\
\midrule
\textbf{KAN-$D_1$}  & \underline{4.3977} & 1.2254 & 5.5063 & \underline{0.9306} & 0.0144 & 72 \\
\textbf{KAN-$D_2$} & 4.4970 & 1.2280 & 5.5008 & 0.9291 & 0.0145 & 60 \\
\textbf{KAN-$D_3$}  & \textbf{4.3957} & {1.2219} & 5.5263 & \textbf{0.9307} & \textbf{0.0142} & 60 \\
\bottomrule
\end{tabular}
\end{table}

\begin{table}[p]
\centering
\caption{Forecasting accuracy comparison of various models (\emph{Period 2}). \textbf{Bold} and \underline{underline} values indicate the best and second-best performances, respectively.}
\label{tab:performance_comparison_p2}
\setlength{\tabcolsep}{5pt}
\renewcommand{\arraystretch}{1.2}
\begin{tabular}{lrrrrrr}
\toprule
\textbf{Model} & \textbf{MSE} & \textbf{MAE} & \textbf{MAPE} & $\bm{R^2}$ & \textbf{QLIKE} & \textbf{\# params} \\
\midrule
Forward filling & 5.2130 & 1.3274 & 5.2876 & 0.9266 & \textbf{0.0124} & - \\
ARMA(1, 1)     & 7.6826 & 1.7531 & 7.1646 & 0.8919 & 0.4931 & 4 \\
ARIMA(1, 1, 1) & 7.9619 & 1.7741 & 7.2251 & 0.8879 & 0.4945 & 4 \\
HAR(3)      & 5.1215 & 1.3029 & 5.2172 & 0.9279 & 0.0125 & 4 \\
HAR(4)      & 5.1222 & 1.3029 & 5.2172 & 0.9279 & 0.0125 & 5 \\
\midrule
MLP-$D_1$ & \textbf{5.0205} & 1.2938 & 5.1739 & \textbf{0.9293} & 0.0130 & 20901 \\
MLP-$D_2$         & 5.4638 & 1.3046 & \underline{5.1737} & 0.9231 & 0.0131 & 20801 \\
MLP-$D_3$          & 5.2307 & \textbf{1.2706} & \textbf{5.0659} & 0.9264 & 0.0130 & 20801 \\
LSTM-$D_1$  & 5.4018 & 1.3172 & 5.2062 & 0.9240 & 0.0133 & 63841 \\
LSTM-$D_2$        & 5.5741 & 1.3694 & 5.3861 & 0.9215 & 0.0134 & 63561 \\
LSTM-$D_3$         & 5.3274 & 1.3890 & 5.5304 & 0.9250 & {0.0129} & 63561 \\
\midrule
\textbf{KAN-$D_1$} & \underline{5.0711} & 1.3051 & 5.2162 & \underline{0.9286} & \underline{0.0125} & 72 \\
\textbf{KAN-$D_2$}& 5.2314 & 1.3155 & 5.2499 & 0.9264 & 0.0127 & 60 \\
\textbf{KAN-$D_3$} & 5.0755 & \underline{1.3003} & 5.2037 & \underline{0.9286} & \textbf{0.0124} & 60 \\
\bottomrule
\end{tabular}
\end{table}

\begin{table}[p]
\centering
\caption{Forecasting accuracy comparison of various models (\emph{Period 3}). \textbf{Bold} and \underline{underline} values indicate the best and second-best performances, respectively.}
\label{tab:performance_comparison_p3}
\setlength{\tabcolsep}{5pt}
\renewcommand{\arraystretch}{1.2}
\begin{tabular}{lrrrrrr}
\toprule
\textbf{Model} & \textbf{MSE} & \textbf{MAE} & \textbf{MAPE} & $\bm{R^2}$ & \textbf{QLIKE} & \textbf{\# params} \\
\midrule
Forward filling & 1.9713 & 1.0065 & 4.5349 & 0.9399 & \underline{0.0074} & - \\
ARMA(1, 1)     & 3.7467 & 1.4055 & 6.3387 & 0.8858 & 0.3298 & 4 \\
ARIMA(1, 1, 1) & 3.7772 & 1.4257 & 6.4268 & 0.8849 & 0.3318 & 4 \\
HAR(3)      & 1.9582 & 0.9939 & 4.5057 & 0.9403 & \underline{0.0074} & 4 \\
HAR(4)      & {1.9579} & 0.9938 & 4.5057 & \underline{0.9403} & \underline{0.0074} & 5 \\
\midrule
MLP-$D_1$ & 1.9846 & \underline{0.9869} & \underline{4.4459} & 0.9395 & 0.0076 & 20901 \\
MLP-$D_2$         & 1.9902 & \textbf{0.9844} & \textbf{4.4250} & 0.9393 & 0.0076 & 20801 \\
MLP-$D_3$          & 2.0092 & 1.0015 & 4.5294 & 0.9388 & 0.0076 & 20801 \\
LSTM-$D_1$ & 1.9884 & 1.0167 & 4.6120 & 0.9394 & 0.0075 & 63841 \\
LSTM-$D_2$        & 2.0544 & 1.0143 & 4.5548 & 0.9374 & 0.0076 & 63561 \\
LSTM-$D_3$         & \textbf{1.9333} & 0.9961 & 4.5186 & \textbf{0.9411} & \textbf{0.0073} & 63561 \\
\midrule
\textbf{KAN-$D_1$} & 1.9702 & 0.9968 & 4.5112 & 0.9400 & 0.0075 & 72 \\
\textbf{KAN-$D_2$}& 1.9955 & 1.0096 & 4.5586 & 0.9392 & 0.0076 & 60 \\
\textbf{KAN-$D_3$} & \underline{1.9575} & {0.9934} & 4.5012 & \underline{0.9403} & \underline{0.0074} & 60 \\
\bottomrule
\end{tabular}
\end{table}

Tables \ref{tab:performance_comparison_p1}--\ref{tab:performance_comparison_p3} summarize the out-of-sample forecasting accuracy for KAN and various benchmark methods.
Across the three \emph{Period}s, KAN consistently delivers competitive results, often achieving the best or second-best performances.
Although the naive forward filling method performs surprisingly well$-$likely owing to the strong persistence inherent in the VIX$-$ARMA(1,1) and ARIMA(1,1,1) reveal the limitations of a purely linear framework devoid of longer-range dependence or multiple timescale averages.
Meanwhile, the HAR(3) and HAR(4) models, which assume a linear additive structure over short-, medium-, long-, and very-long-term averages, generally achieve decent results.

Regarding neural-network models, MLP and LSTM often demonstrate strong predictive power.
However, they require approximately 300 or 1,000 times more parameters compared to KAN while offering comparable accuracy.
Moreover, a closer look at the metrics suggests that KAN tends to excel in MSE rather than MAE or MAPE.
This indicates that KAN produces fewer extreme errors or outliers.
Remarkably, KAN’s parsimonious design consistently achieves performance on par with these more parameter-intensive MLP-based networks, underscoring its robustness and stability in noisy financial time-series data.

Turning to the different dataset configurations, \emph{Dataset 3} typically yields the best results for KAN, MLP, LSTM, and HAR models alike.
Notably, KAN–$D_3$ demonstrates particularly strong accuracy in all three periods.
This suggests that effectively capturing key characteristics of the VIX, such as mean-reversion and high-persistence, leads to reliable forecasting results. 
Models using \emph{Dataset 1} also perform reasonably well.
The decent results when using \emph{Dataset 1} are presumably because consecutive lags capture crucial short-term dependencies (e.g., volatility clustering, momentum).
By contrast, \emph{Dataset 2} often fails to provide significant multi-horizon features.

Taken together, these findings confirm that KAN offers a robust and interpretable alternative to both classical time-series approaches and parameter-intensive MLP-based neural networks.
This illustrates that a parsimonious structure can yield high forecasting accuracy in noisy financial time-series such as the VIX.

\section{Conclusion} \label{sec: conclusion}

MLP-based neural networks have exhibited remarkable forecasting performances in numerous studies, but their black-box nature often restricts their practical use in financial time-series forecasting such as the VIX.
In this study, we employ Kolmogorov–Arnold Networks (KANs) to address these limitations.
Each edge in KAN extracts important features from the data and describes them with symbolic functions, enabling the output of the network to be expressed as a closed-form in terms of explanatory variables.
This interpretability allows us to gain valuable insights into the VIX's inherent characteristics such as mean reversion and the leverage effect.
Furthermore, our empirical analysis across multiple periods and datasets indicates that KAN achieves robust and competitive accuracy compared to MLPs and LSTMs while requiring significantly fewer parameters and maintaining interpretability. 
These findings suggest that KAN can serve as a powerful and parsimonious framework for analyzing noisy financial time-series data, including the VIX, and may be extended to broader contexts where clear, data-driven insights are essential.

% % The VIX, often referred to as the “fear index” is an important measure of the forward-looking volatility of the market expectations for the S\&P 500. 
% The VIX, often referred to as the ``fear index" represents the market's expected volatility of S\&P 500 options over the next 30 days and remains a pivotal topic in financial research due to its significance in understanding market dynamics.
% In this study, we employ Kolmogorov-Arnold Networks (KANs) which address the limitations of MLP-based models, black-box nature, by allowing activation functions to explicitly represent effects of input variables thus enhancing interpretability. 
% By leveraging the interpretability of KANs, we not only predicted the VIX but also gained valuable insights into its mean-reverting nature and the leverage effect.
% Furthermore, KANs demonstrated efficiency and practicality by achieving competitive performance comparable to that of MLPs with significantly fewer parameters, underscoring their potential as a powerful alternative for financial modeling.

\appendix{}
\setcounter{table}{0}
\setcounter{figure}{0}

\section{Training results for deep and wide KAN} \label{appendix: with deep depth and wide width}

\begin{figure}[!htb]
    \centering
    %---------------- Row 1 (Top row of 3 images) ----------------%
    \begin{subfigure}[t]{0.33\textwidth}
        \centering
        \includegraphics[width=\linewidth]{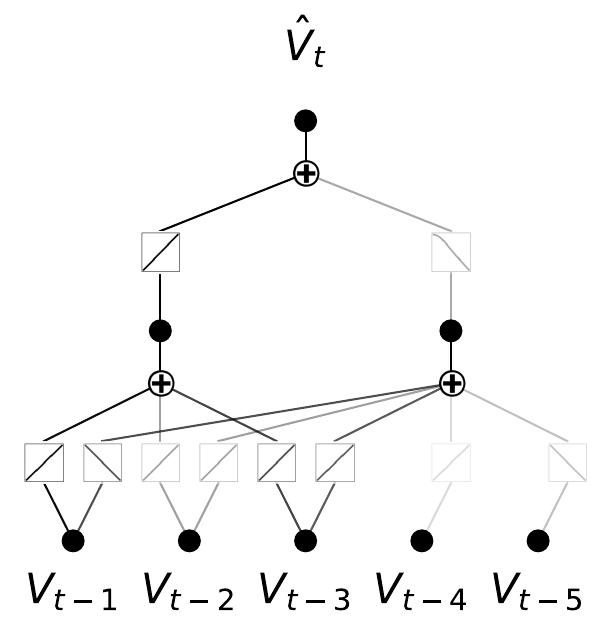}
        \caption{$D_1P_3$ with two layers}
    \end{subfigure}
    \hfill
    \begin{subfigure}[t]{0.33\textwidth}
        \centering
        \includegraphics[width=\linewidth]{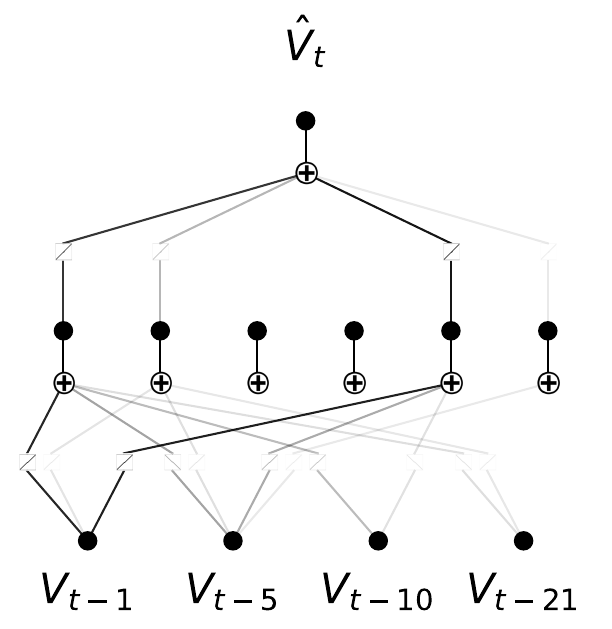}
        \caption{$D_2P_3$ with two layers}
    \end{subfigure}
    \hfill
    \begin{subfigure}[t]{0.33\textwidth}
        \centering
        \includegraphics[width=\linewidth]{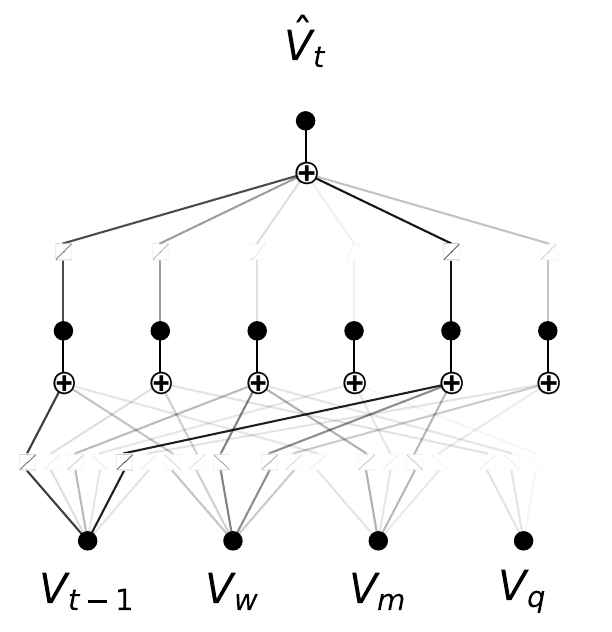}
        \caption{$D_3P_3$ with two layers}
    \end{subfigure}

    %---------------- Row 2 (Bottom row of 3 images) ----------------%
    \vspace{0.5cm} % Optional: Add vertical spacing between rows
    \begin{subfigure}[t]{0.33\textwidth}
        \centering
        \includegraphics[width=\linewidth]{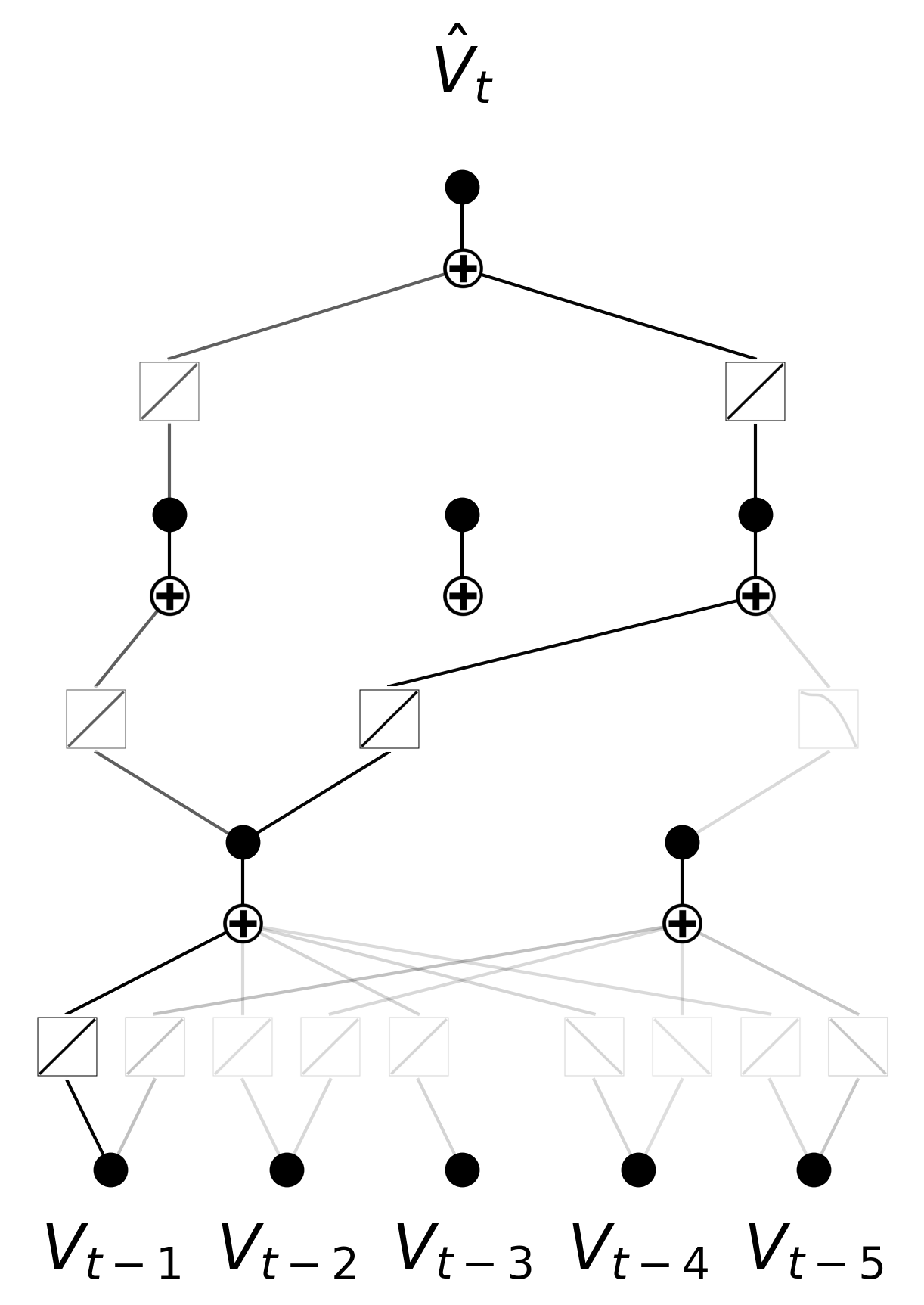}
        \caption{$D_1P_3$ with three layers}
    \end{subfigure}
    \hfill
    \begin{subfigure}[t]{0.33\textwidth}
        \centering
        \includegraphics[width=\linewidth]{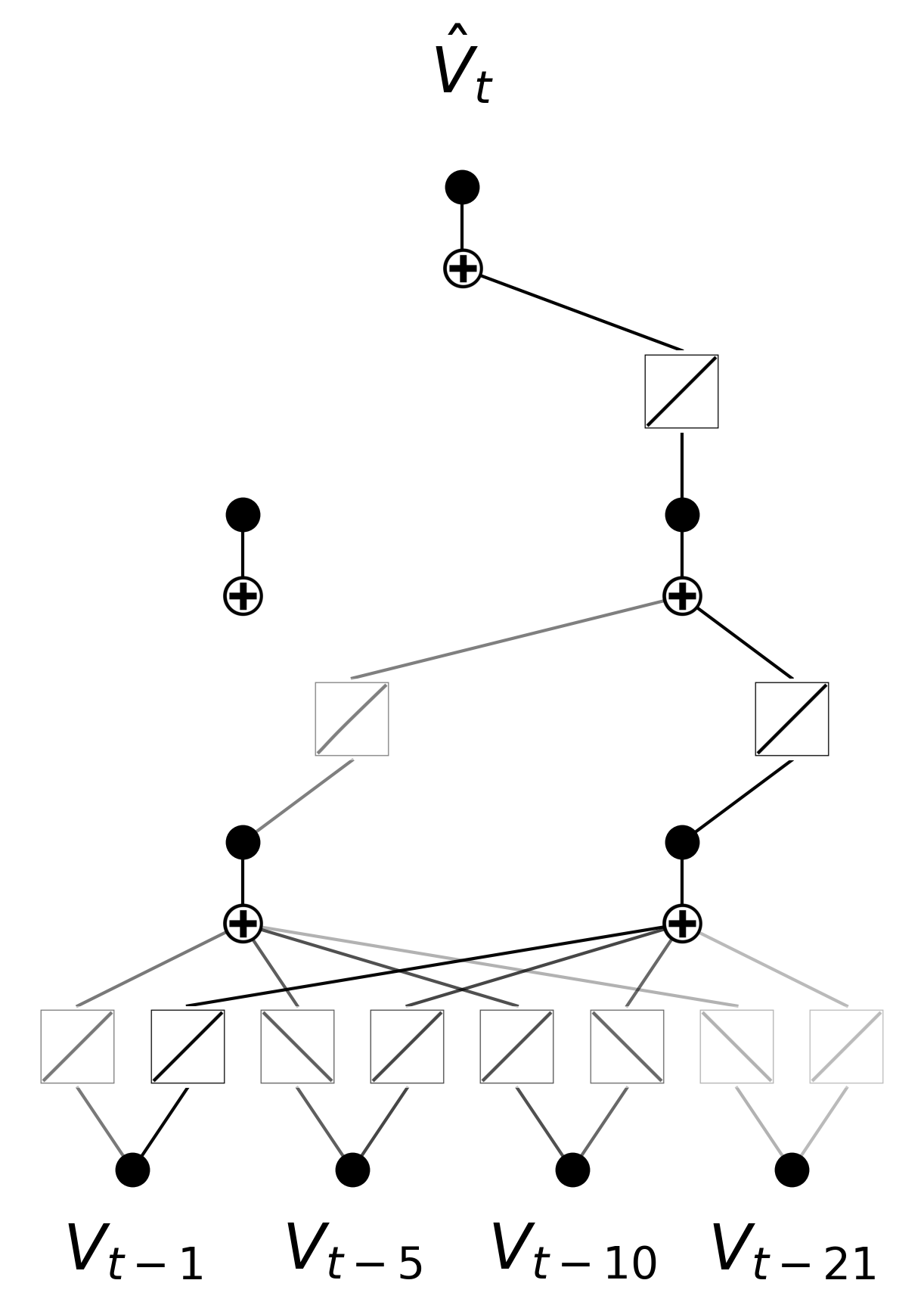}
        \caption{$D_2P_3$ with three layers}
    \end{subfigure}
    \hfill
    \begin{subfigure}[t]{0.33\textwidth}
        \centering
        \includegraphics[width=\linewidth]{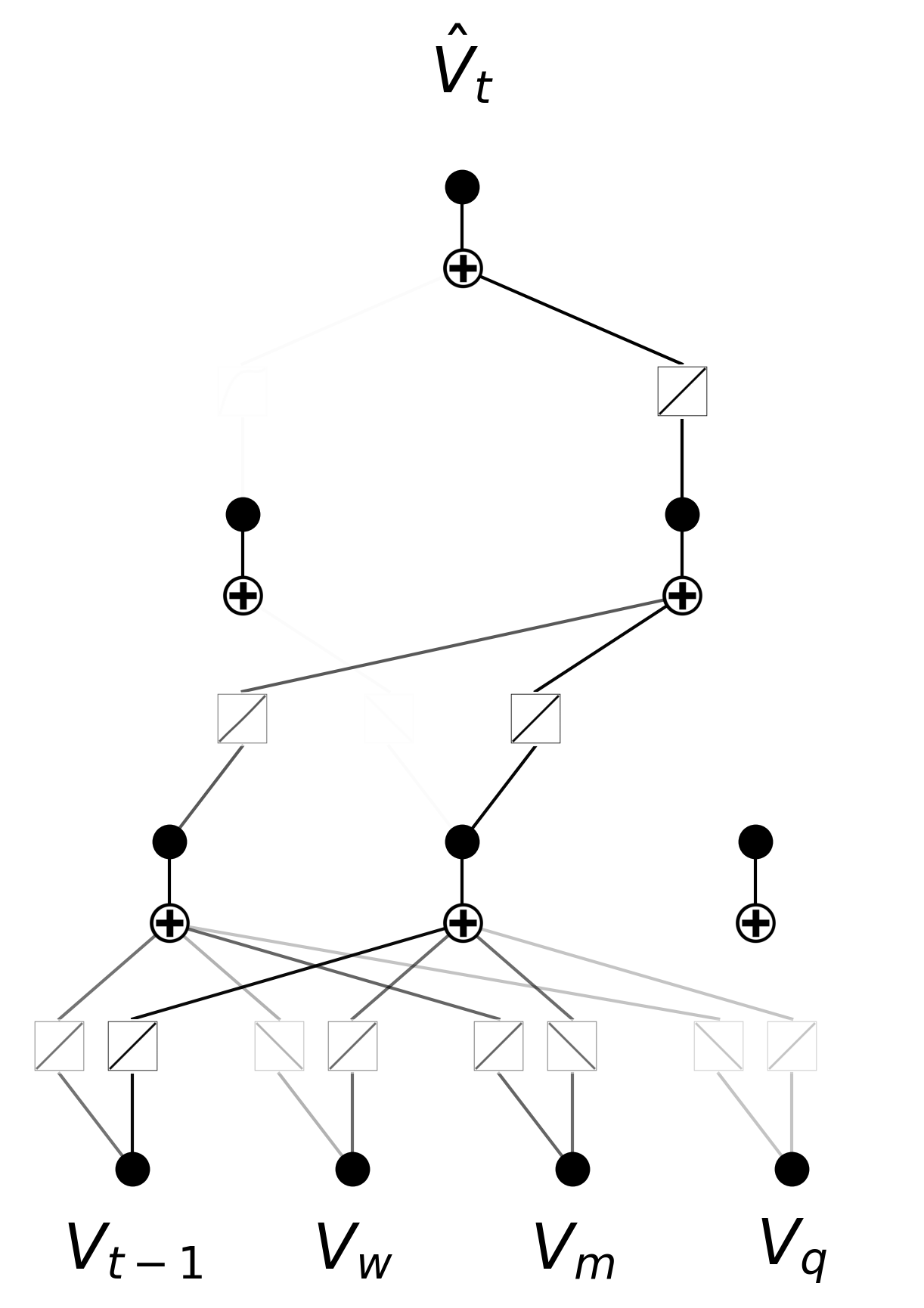}
        \caption{$D_3P_3$ with three layers}
    \end{subfigure}

    \caption{KAN training results for \emph{Datasets 1--3} under \emph{Period 3} with 10 hidden nodes (top row) and with three layers and 4 hidden nodes each (bottom row). The trained activation functions predominantly exhibit a linear form, leading to the KAN model being expressed as a sum of linear terms. This structure is preserved across all \emph{Dataset}s.}
    \label{fig:KAN with large width and depth}
\end{figure}

In this paper, all experiments were conducted with a two-layer KAN using two nodes in the hidden layer. 
This section examines the impact of increasing either the depth or the number of nodes. Specifically, we analyze two cases: (1) a two-layer KAN with 10 hidden nodes, and (2) a three-layer KAN with 4 hidden nodes each. 
The training results are presented in Figure \ref{fig:KAN with large width and depth}. 
While depth and width can be adjusted to various values, the results remain consistent and are therefore omitted.
As shown in the figure, even with increased width or depth, the learned activation functions for each layer predominantly exhibit linear characteristics.
The actual number of active nodes in the figure may appear smaller (e.g., 2 or 6) than the predefined configuration with a larger number of nodes (e.g., 10). 
This is because, during training, unimportant nodes are pruned as part of the optimization process, leaving only significant nodes to contribute to the final model.
All these activation functions are replaced with linear functions $y=x$ or $y=-x$ at the symbolification step, and summed at the node level.
Consequently, the outcomes of each layer become a linear combination of input variables.
Even after passing through multiple layers, the result remains a linear combination.
Thus, we chose the simplest configuration, [$n_i$, 2, 1], where $n_i$ = 4 for \emph{Dataset 1} and $n_i$ = 5 for \emph{Dataset 2} and \emph{Dataset 3}, as the structure of the KAN model.

\section{KAN training result before and after symbolification in \emph{Period 1} and \emph{Period 2}} \label{appendix: symbolification experiment P1 and P2}

\begin{figure}[p]
    \centering
    %---------------- Row j=1 (Train Period 1) ----------------%
    \begin{subfigure}[b]{0.23\textwidth}
        \centering
        \includegraphics[width=\linewidth]{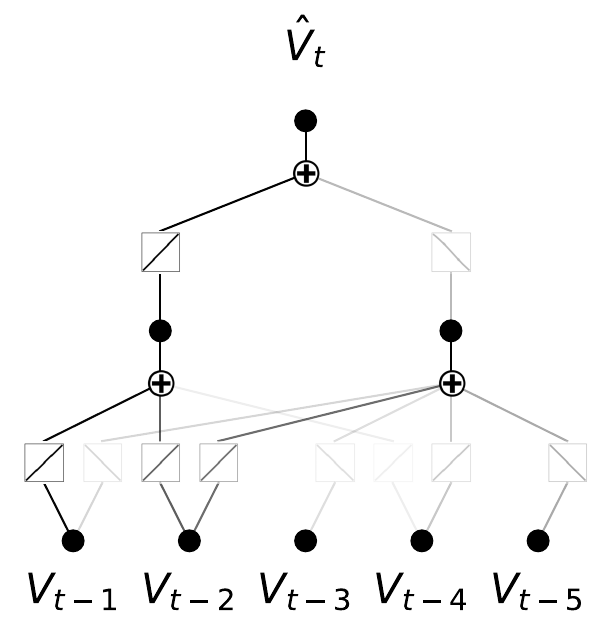}
        \caption{$D_1P_1$}
    \end{subfigure}
    \hspace{0.08\textwidth}
    \begin{subfigure}[b]{0.23\textwidth}
        \centering
        \includegraphics[width=\linewidth]{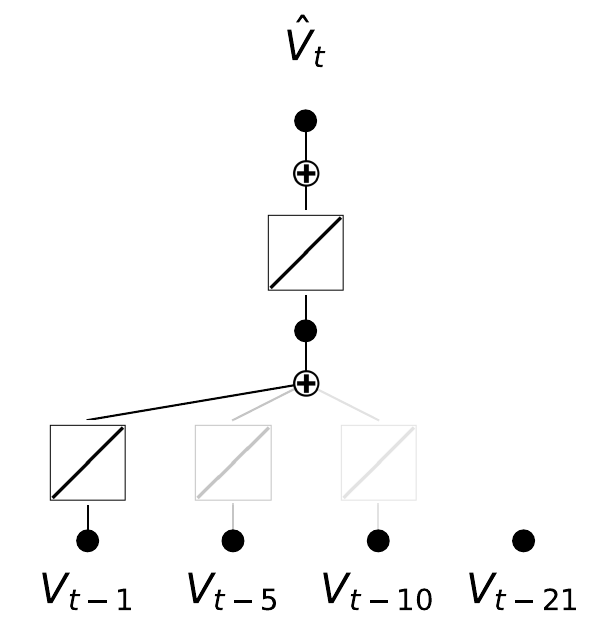}
        \caption{$D_2P_1$}
    \end{subfigure}
    \hspace{0.08\textwidth}    \begin{subfigure}[b]{0.23\textwidth}
        \centering
        \includegraphics[width=\linewidth]{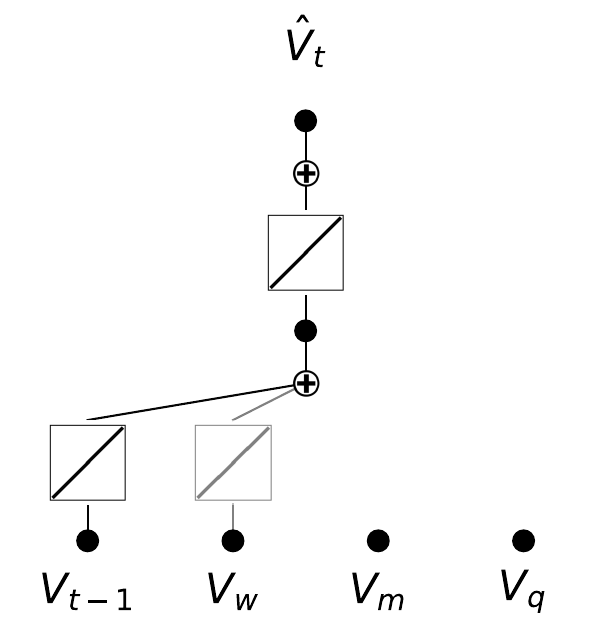}
        \caption{$D_3P_1$}
    \end{subfigure}
    
    \vskip\baselineskip

    %---------------- Row j=2 (Train Period 2) ----------------%
    \begin{subfigure}[b]{0.23\textwidth}
        \centering
        \includegraphics[width=\linewidth]{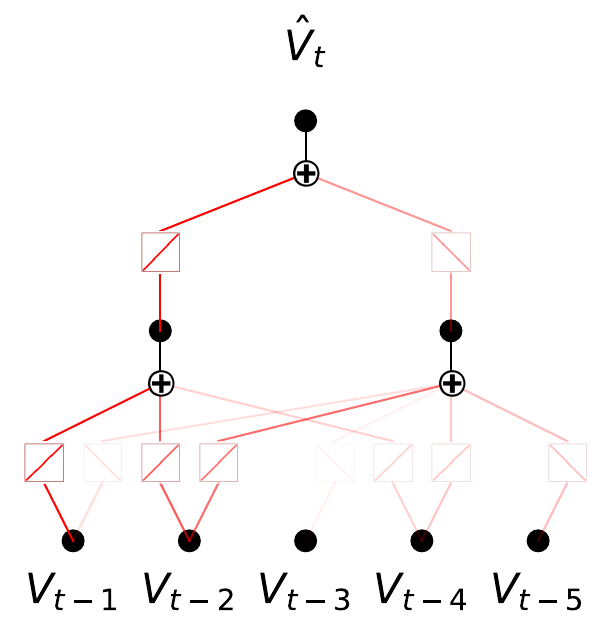}
        \caption{$D_1P_1$}
    \end{subfigure}
    \hspace{0.08\textwidth}    \begin{subfigure}[b]{0.23\textwidth}
        \centering
        \includegraphics[width=\linewidth]{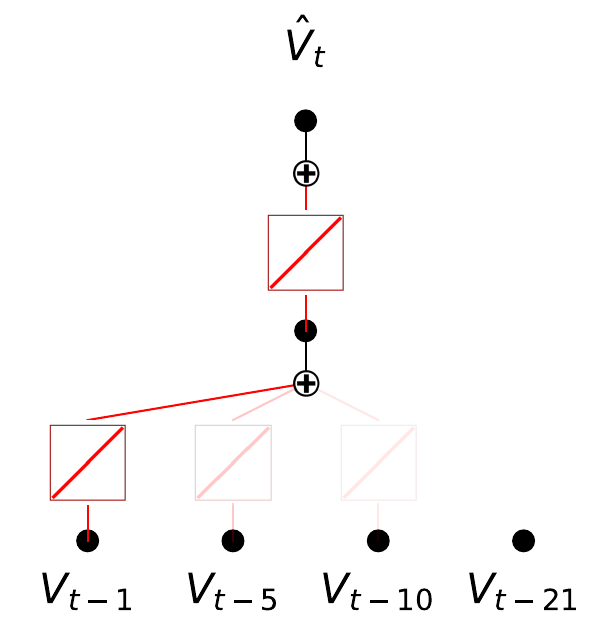}
        \caption{$D_2P_1$}
    \end{subfigure}
    \hspace{0.08\textwidth}    \begin{subfigure}[b]{0.23\textwidth}
        \centering
        \includegraphics[width=\linewidth]{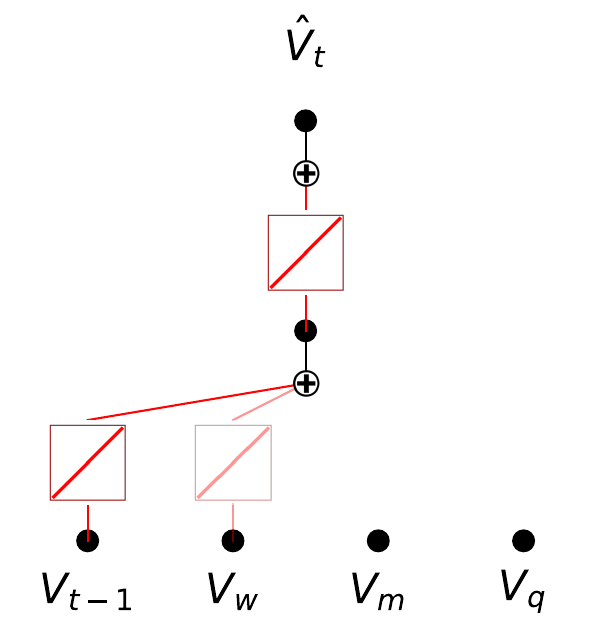}
        \caption{$D_3P_1$}
    \end{subfigure}

    \caption{KAN training results for \emph{Datasets 1--3} under \emph{Period 1} {before} (top row) and {after} (bottom row) symbolification. Black activation functions indicate the trained B-spline-based structure, whereas red activation functions depict the replaced symbolic functions. Vivid edges represent stronger effects, while fainter edges mean weaker ones.}
    \label{fig: kan training results period 1}
\end{figure}

\begin{figure}[p]
    \centering
    %---------------- Row j=1 (Train Period 1) ----------------%
    \begin{subfigure}[b]{0.23\textwidth}
        \centering
        \includegraphics[width=\linewidth]{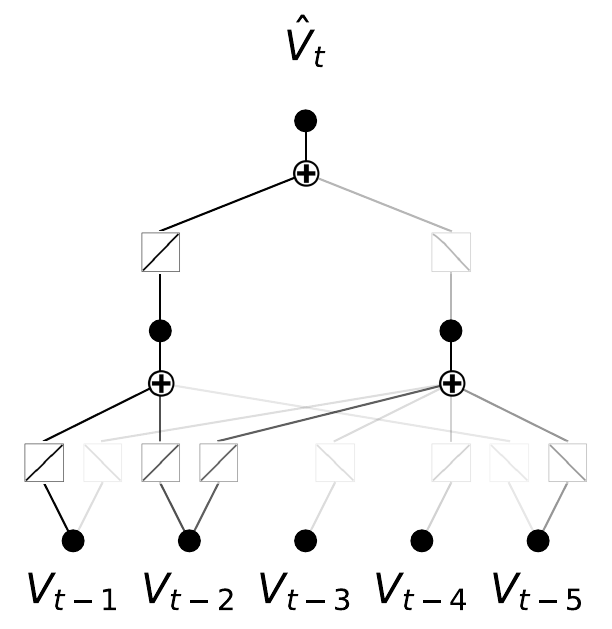}
        \caption{$D_1P_2$}
    \end{subfigure}
    \hspace{0.08\textwidth}    \begin{subfigure}[b]{0.23\textwidth}
        \centering
        \includegraphics[width=\linewidth]{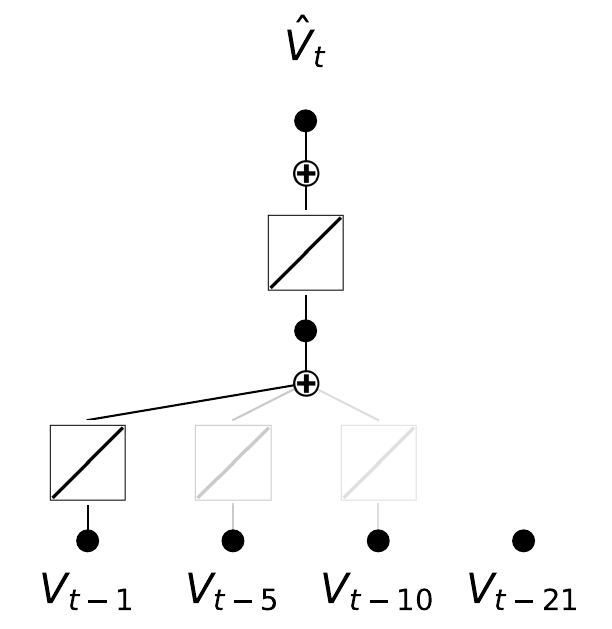}
        \caption{$D_2P_2$}
    \end{subfigure}
    \hspace{0.08\textwidth}    \begin{subfigure}[b]{0.23\textwidth}
        \centering
        \includegraphics[width=\linewidth]{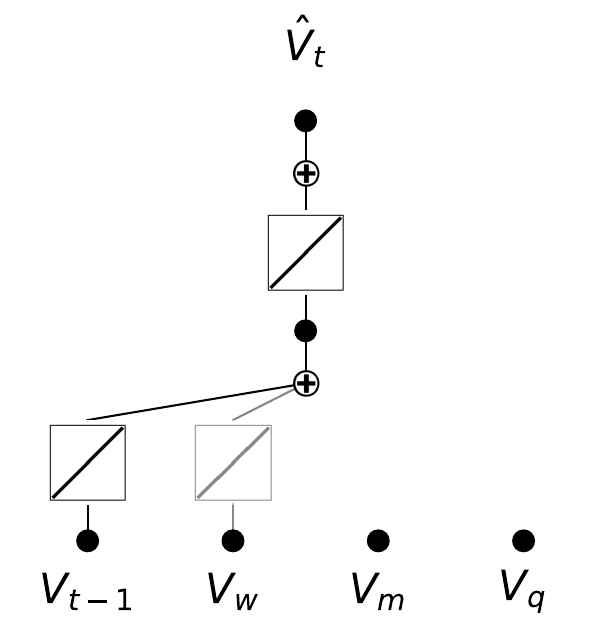}
        \caption{$D_3P_2$}
    \end{subfigure}
    
    \vskip\baselineskip

    %---------------- Row j=2 (Train Period 2) ----------------%
    \begin{subfigure}[b]{0.23\textwidth}
        \centering
        \includegraphics[width=\linewidth]{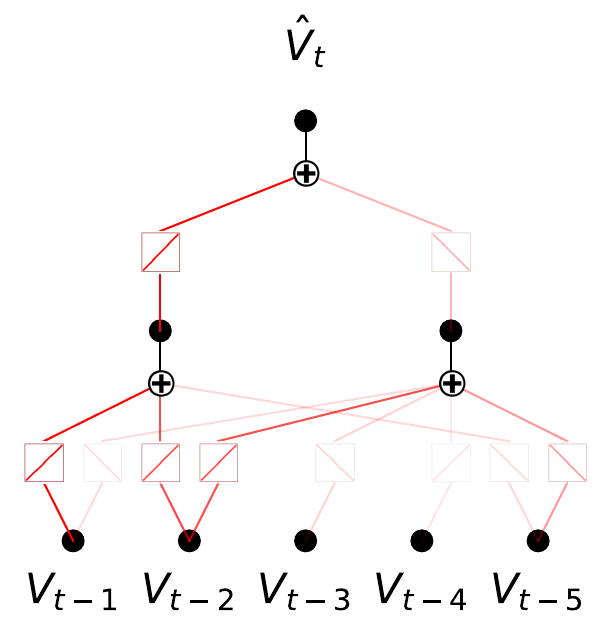}
        \caption{$D_1P_2$}
    \end{subfigure}
    \hspace{0.08\textwidth}    \begin{subfigure}[b]{0.23\textwidth}
        \centering
        \includegraphics[width=\linewidth]{figs/dataset_2_train_period_2.pdf}
        \caption{$D_2P_2$}
    \end{subfigure}
    \hspace{0.08\textwidth}    \begin{subfigure}[b]{0.23\textwidth}
        \centering
        \includegraphics[width=\linewidth]{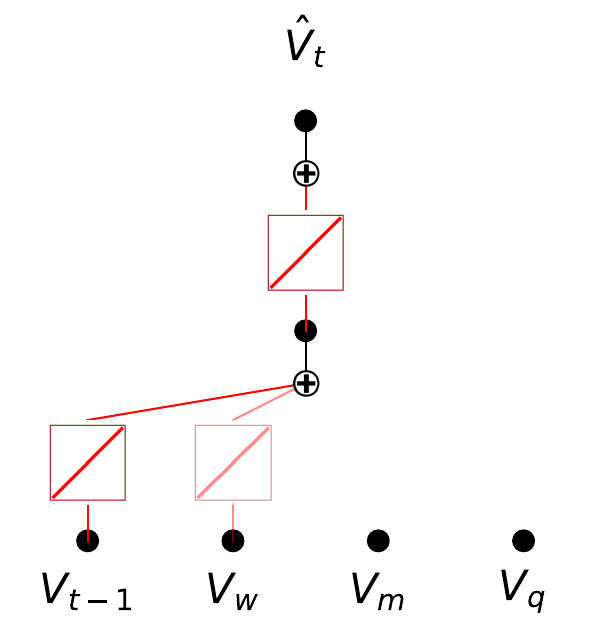}
        \caption{$D_3P_2$}
    \end{subfigure}

    \caption{KAN training results for \emph{Datasets 1--3} under \emph{Period 2} {before} (top row) and {after} (bottom row) symbolification. Black activation functions indicate the trained B-spline-based structure, whereas red activation functions depict the replaced symbolic functions. Vivid edges represent stronger effects, while fainter edges mean weaker ones.}
    \label{fig: kan training results period 2}
\end{figure}

In Section \ref{sec: kan result analysis}, only the training results of KANs in \emph{Period 3} are presented. 
The results in \emph{Period 1} and \emph{Period 2} are illustrated in Figure \ref{fig: kan training results period 1} and Figure \ref{fig: kan training results period 2}, respectively.
As shown in the figures, it is evident that the trained activation functions before symbolification exhibit a linear form across all \emph{Dataset}s and \emph{Period}s.

\section*{Acknowledgement}
The research of Sungchul Lee was supported by the National Research Foundation of Korea grant funded by the Korea government (Grant No. RS-2024-00456958).

\section*{CRediT authorship contribution statement}
\textbf{So-Yoon Cho}: Conceptualization, Software, Writing - Original Draft. \textbf{Sungchul Lee}: Writing - Review \& Editing, Validation. \textbf{Hyun-Gyoon Kim}: Conceptualization, Methodology, Writing - Original Draft, Supervision.

\section*{Data availability}
The data that support the findings of this study are available from DataGuide. Restrictions apply to the availability of these data, which were used under license for this study. Data are available from Yonsei University with the permission of DataGuide.

\section*{Declaration of competing interests}
The authors declare that they have no known competing financial interests or personal relationships that could have appeared to influence the work reported in this paper.

\section*{Declaration of generative AI and AI-assisted technologies in the writing process}
During the preparation of this work the authors used ChatGPT in order to improve the grammatical quality and readability. After using this service, the authors reviewed and edited the content as needed and take full responsibility for the content of the publication.

\bibliography{KAN_bib}
% \bibliography{main}

\end{document}